\documentclass{article}
\PassOptionsToPackage{numbers,sort&compress}{natbib}

\usepackage[final,main]{neurips_2026}

\usepackage[utf8]{inputenc}
\usepackage[T1]{fontenc}
\usepackage{hyperref}
\usepackage{url}
\usepackage{booktabs}
\usepackage{array}
\usepackage{amsmath}
\usepackage{amsfonts}
\usepackage{amssymb}
\usepackage{graphicx}
\usepackage{xcolor}
\usepackage{multirow}
\usepackage{subcaption}
\usepackage{caption}
\usepackage{xspace}
\setcitestyle{numbers,sort&compress}
\usepackage{wrapfig}

\newcommand{\method}{WovenAnchor Matcher\xspace}
\newcommand{\methodshort}{WAM\xspace}
\newcommand{\wm}{WovenMamba\xspace}
\newcommand{\aca}{AnchorCrossAttention\xspace}

\title{WovenAnchor Matcher: Specialized Intra- and Inter-Image Context Modeling for 
Feature Matching}

\author{
    Zhiyang Li$^{1}$ \quad Ruijiang Jin$^{1}$ \quad Thibaut Klenke$^{1}$ \quad
    \textbf{Yusuke Sekikawa}$^{2}$ \quad \textbf{Nakamasa Inoue}$^{1}$ \\
    $^{1}$Institute of Science Tokyo, Tokyo, Japan \quad 
    $^{2}$Denso IT Laboratory, Tokyo, Japan \\
    {\small \texttt{li.z.6a09@m.isct.ac.jp}}
}

\begin{document}
\maketitle

\begin{abstract}
Semi-dense local feature matching commonly aggregates contextual information before coarse-to-fine correspondence estimation. 
However, intra-image and inter-image aggregation serve different roles: the former should propagate spatial evidence within each image while retaining structured two-dimensional dependencies, whereas the latter should gather information from plausible corresponding regions while limiting noise from unrelated locations. 
We propose \textbf{WovenAnchor Matcher (WAM)}, a semi-dense matching framework based on \emph{role-specialized context aggregation}. 
WAM introduces two complementary operators. 
\textbf{WovenMamba} performs intra-image aggregation through horizontal-then-vertical state-space propagation, allowing vertical updates to operate on horizontally contextualized features and thereby encouraging a structured two-dimensional receptive-field bias. 
\textbf{AnchorCrossAttention} performs inter-image aggregation by using cross-attention to gather context from plausible corresponding regions across the image pair. It performs this exchange over local representative anchors that aggregate feature evidence within each window.
On MegaDepth, WAM achieves 66.1 pose AUC@5$^\circ$, improving on the JamMa baseline by 2.0 AUC points under the same evaluation setup.
Ablations show that removing or replacing either specialized component reduces accuracy, providing evidence that intra-image propagation and inter-image retrieval benefit from role-specific aggregation designs.
Code and pretrained weights are available at \url{https://github.com/mmaiLab/WovenAnchorMatcher}.
\end{abstract}

\section{Introduction}

Local feature matching is a core component of visual localization, structure-from-motion, SLAM, and image alignment \citep{sattler2018benchmarking,sarlin2019coarse,schoenberger2016sfm,murartal2015orbslam,hartley2004multiple}. 
Modern semi-dense matchers estimate many correspondences from learned feature maps and commonly follow a coarse-to-fine pipeline \citep{sun21loftr,lu2025jamma,Li2025EDM}. 
Before coarse matching, a context aggregation block updates descriptors using information from both the same image and the paired image.

These two sources of context are not interchangeable. 
Intra-image aggregation should propagate spatial evidence while respecting the image's two-dimensional layout. 
Inter-image aggregation should instead gather evidence from plausible corresponding regions and avoid contaminating descriptors with visually similar but unrelated content. 
This matters because dense affinities can be unreliable under repetition, viewpoint changes, and local ambiguity.

Existing matchers only partially encode this asymmetry. 
LoFTR-style methods explicitly separate self-attention and cross-attention, but both rely on attention-based token interaction as a generic aggregation primitive \citep{vaswani2017attention,sun21loftr,wang22matchformer,chen22aspanformer,wang24eloftr}. 
JamMa improves efficiency by replacing attention-heavy aggregation with Mamba-based sequence modeling, but its JointMamba block applies a closely related sequence-modeling bias to within-image propagation and cross-image interaction \citep{lu2025jamma}. 
These designs are effective, but do not fully tailor their aggregation structures to the distinct requirements of intra-image propagation and inter-image retrieval.

\begin{wrapfigure}[18]{r}{0.5\linewidth}
    \centering
    \includegraphics[width=\linewidth]{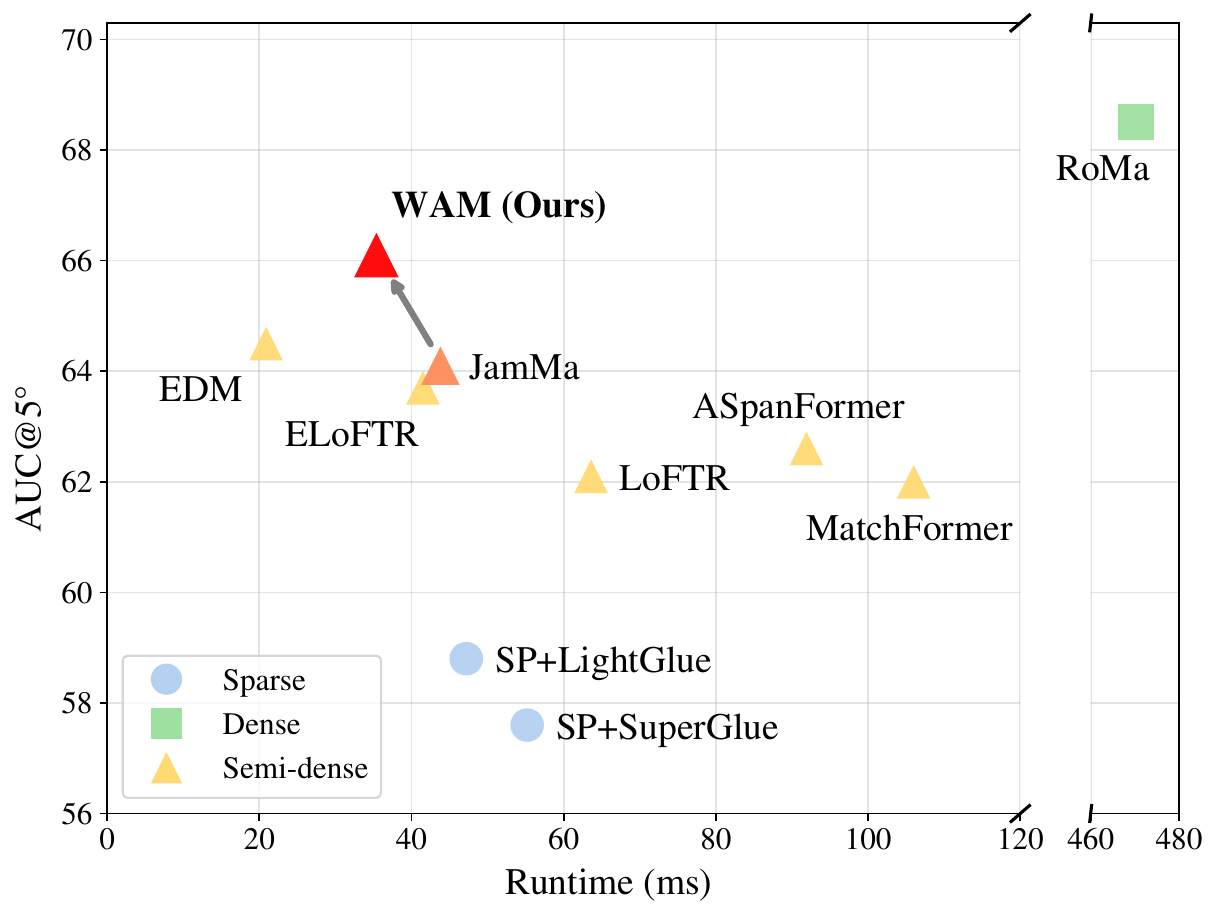}
    \caption{\textbf{Relative pose AUC@5$^\circ$ vs. runtime on MegaDepth.} Runtime is measured on a single NVIDIA H100 GPU.}
    \label{fig:tradeoff}
\end{wrapfigure}

We propose \method{} (\methodshort), a semi-dense matcher that makes this operator-level asymmetry explicit. 
WAM keeps the standard coarse-to-fine matching framework but replaces the generic context block with a role-specialized \emph{WovenAnchorModule}. 
For intra-image aggregation, \wm{} applies horizontal state-space propagation followed by vertical propagation. 
Since the vertical stage operates on horizontally contextualized features, the operator encourages composed two-dimensional dependencies rather than an additive combination of independent directional scans. 
We analyze this bias with a simplified linear model on a finite grid. This structured receptive field is important because intra-image context should preserve where supporting evidence appears, not only whether similar evidence exists somewhere in the image.

For inter-image aggregation, \aca{} uses cross-attention to gather context from plausible corresponding regions, but performs this exchange over local representative anchors instead of dense point-wise tokens \citep{bezdek1981fuzzy,Arandjelovic2016NetVLAD,lee2019set}. 
Multiple anchors are formed within each local neighborhood, allowing different structures in the same area to receive different cross-image updates. 
This design aims to reduce reliance on ambiguous token-level affinities by aggregating local evidence before cross-image retrieval.

Experiments support the value of this role-specialized design. On MegaDepth, WAM improves over JamMa under our setup, increasing pose AUC@5$^\circ$ from 64.1 to 66.1. Component-removal and operator-replacement ablations show that removing or substituting either \wm{} or \aca{} lowers accuracy.
Shared/reversed-role experiments further support the proposed assignment of \wm{} to intra-image propagation and \aca{} to inter-image retrieval.


Our contributions are threefold. 
First, we formulate semi-dense context aggregation as two roles with different inductive biases: spatially structured intra-image propagation and correspondence-aware inter-image retrieval. Second, we introduce \wm{}, a horizontal-then-vertical SSM composition whose simplified receptive field exhibits direction- and distance-dependent two-dimensional support. Third, we introduce \aca{}, a cross-attention operator that retrieves cross-image context through locally grouped anchors.

\section{Related work}

\paragraph{Local feature matching.}
Feature matching methods can be grouped into sparse, dense, and semi-dense approaches. Sparse pipelines detect keypoints and match descriptors \citep{Lowe2004SIFT,Rublee2011ORB}, with representative learned systems including SuperPoint with SuperGlue and LightGlue \citep{detone18superpoint,sarlin20superglue,sarlin23lightglue}. Dense matchers such as DKM and RoMa estimate correspondences over denser grids and can achieve strong accuracy at higher computational cost \citep{edstedt23dkm,edstedt24roma}. Semi-dense matchers seek a balance by learning dense coarse features and refining selected correspondences; representative examples include LoFTR, JamMa, and EDM \citep{sun21loftr,lu2025jamma,Li2025EDM}. WAM belongs to this semi-dense family and is most directly related to JamMa in its overall coarse-to-fine framework.

\paragraph{Context aggregation for semi-dense matching.}
LoFTR introduced detector-free matching with self-attention and cross-attention over coarse feature maps \citep{vaswani2017attention,sun21loftr}. Later methods improve attention structure, feature interaction, or efficiency while retaining attention-based aggregation \citep{wang22matchformer,chen22aspanformer,wang24eloftr}. JamMa replaces attention-heavy aggregation with a Mamba-based module for efficient long-range modeling \citep{lu2025jamma}. WAM follows this JamMa-like coarse-to-fine setting, but differs mainly in context aggregation. Instead of using a shared mechanism for both roles, WAM separates them into structured spatial propagation within each image and correspondence-aware aggregation across images.

\paragraph{State-space models and compact cross-image interaction.}
State-space sequence models such as Mamba provide an efficient alternative to attention for long-range modeling \citep{Gu2024Mamba,dao2024mamba2}, and recent vision variants adapt scan-based SSMs to image grids \citep{Zhu24VisionMamba,liu2024vmamba}. For image features, the scan order determines the spatial inductive bias; WovenMamba differs from scan-and-sum designs by composing the two scan orders rather than summing independent directional responses. For inter-image aggregation, dense cross-attention can mix information from non-corresponding regions, especially when noise or visual ambiguity increases their apparent similarity. This motivates compact local representations for more correspondence-aware cross-image aggregation.

\section{Method}
\label{sec:method}

\subsection{Overview}

Given an image pair $(I^0,I^1)$, the backbone extracts coarse feature maps $\mathbf{F}_c^0,\mathbf{F}_c^1\in\mathbb{R}^{H_c\times W_c \times C_c}$ and fine feature maps $\mathbf{F}_f^0,\mathbf{F}_f^1$. WAM updates the coarse features with \textbf{WovenAnchorModule} stages. Each stage first applies \wm independently to the two images for structured intra-image propagation, then applies \aca to aggregate cross-image context around likely correspondences, followed by a feed-forward block. The updated coarse features are used for coarse matching, while the fine features are used for standard local refinement.

\begin{figure*}[t]
    \centering
    \begin{subfigure}[t]{0.4\linewidth}
        \centering
        \includegraphics[width=\linewidth]{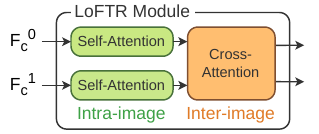}
        \caption{LoFTR context aggregation.}
    \end{subfigure}
    \hspace{0.04\linewidth}
    \begin{subfigure}[t]{0.4\linewidth}
        \centering
        \includegraphics[width=\linewidth]{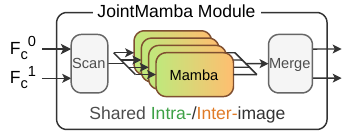}
        \caption{JamMa shared context aggregation.}
    \end{subfigure}

    \vspace{0.8em}

    \begin{subfigure}[t]{1.0\linewidth}
        \centering
        \includegraphics[width=\linewidth]{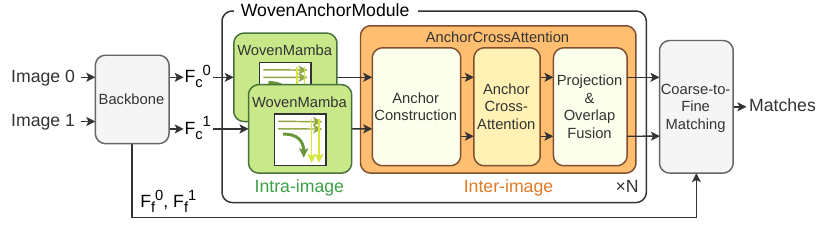}
        \caption{WAM specialized context aggregation.}
    \end{subfigure}
    \caption{\textbf{Context aggregation designs.}
    LoFTR uses separate self-/cross-attention, JamMa uses a shared JointMamba design, and WAM explicitly separates intra-image and inter-image aggregation into WovenMamba and AnchorCrossAttention. The diagrams are schematic summaries of the coarse context aggregation blocks rather than layer-by-layer implementations.}
    \label{fig:architecture}
\end{figure*}

Figure~\ref{fig:architecture} summarizes the architectural distinction. LoFTR uses self-attention and cross-attention, and JamMa uses a shared JointMamba aggregation module, whereas WAM separates intra-image and inter-image aggregation into specialized components. We first describe WovenMamba and the receptive-field analysis that motivates its design, and then present AnchorCrossAttention.

\subsection{WovenMamba for intra-image aggregation}
\label{sec:wovenmamba}
Since local appearance alone may be insufficient to distinguish visually similar features, especially in repetitive patterns or locally ambiguous regions, WovenMamba enriches each descriptor with surrounding two-dimensional spatial context. To preserve the directional and distance structure of this context, it uses horizontal-then-vertical state-space propagation rather than scan-and-sum aggregation of independent directional scans.

Let $\mathbf{F}\in\mathbb{R}^{H\times W\times C}$ denote the input feature map, with the batch dimension omitted for clarity. WovenMamba first maps the normalized input into a propagation branch $\mathbf{X}$ and a gate branch $\mathbf{Z}$:
\begin{equation}
    (\mathbf{X},\mathbf{Z})
    =
    \operatorname{Split}\!\left(
    \operatorname{Linear}_{\mathrm{in}}(\operatorname{Norm}(\mathbf{F}))
    \right).
\end{equation}
Here, $\operatorname{Split}$ divides the projected channels into the propagation and gate branches.

For an order $a\in\{h,v\}$, let $\mathcal{S}_a$ denote a Mamba-style SSM scan \citep{Gu2024Mamba} applied after flattening the grid according to a horizontal or vertical order:
\begin{equation}
    \mathcal{S}_a(\mathbf{G})
    =
    \operatorname{Unscan}_a\!\left(
    \operatorname{SSM}_a(\operatorname{Scan}_a(\mathbf{G}))
    \right).
\end{equation}
Here, $\operatorname{Scan}_h$ and $\operatorname{Scan}_v$ list grid positions in row-major order and column-major order, respectively; $\operatorname{Unscan}_a$ maps the scanned sequence back to the grid.
For a scanned sequence $\{\mathbf{s}_t\}_{t=1}^{L}$, $\operatorname{SSM}_a$ produces $\{\mathbf{o}_t\}_{t=1}^{L}$ through the selective state-space update
\begin{equation}
    \mathbf{h}_t
    =
    \bar{\mathbf{A}}_t \mathbf{h}_{t-1}
    +
    \bar{\mathbf{B}}_t \mathbf{s}_t,
    \qquad
    \mathbf{o}_t
    =
    \mathbf{C}_t \mathbf{h}_t
    +
    \mathbf{D}\mathbf{s}_t .
\end{equation}
Here, $\mathbf{h}_t$ is the recurrent state; $\bar{\mathbf{A}}_t$ and $\bar{\mathbf{B}}_t$ are input-dependent transition and input terms after discretization; $\mathbf{C}_t$ is an input-dependent output term; and $\mathbf{D}$ is a direct skip term.

WovenMamba then computes
\begin{equation}
    \mathbf{O}
    =
    \mathcal{S}_v\!\left(
    \operatorname{Norm}(\mathcal{S}_h(\mathbf{X}))
    \right),
    \qquad
    \mathbf{Y}
    =
    \mathbf{F}
    +
    \operatorname{Linear}_{\mathrm{out}}
    \left(
    \mathbf{O}\odot \operatorname{SiLU}(\mathbf{Z})
    \right).
\end{equation}
Here, $\operatorname{Norm}$ denotes the intermediate normalization between the two scans, $\operatorname{SiLU}$ is the gating nonlinearity, and $\odot$ denotes element-wise multiplication. The key design is the composition $\mathcal{S}_v(\operatorname{Norm}(\mathcal{S}_h(\mathbf{X})))$: the vertical scan operates on horizontally contextualized features, producing a composed two-dimensional dependency pattern instead of an additive combination of independent directional scans.

\begin{figure}[t]
    \centering
    \includegraphics[width=0.9\linewidth]{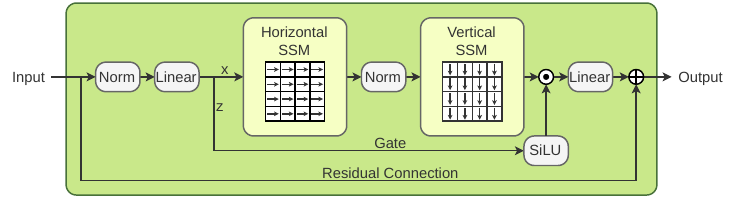}
    \caption{\textbf{WovenMamba.} The input is split into propagation and gate branches. The propagation branch is processed by horizontal-then-vertical SSM propagation, then gated, projected, and added back through a residual connection. 
    The arrows illustrate the dominant horizontal and vertical scan orders; the implementation uses the corresponding row-major and column-major flattenings.}
    \label{fig:wovenmamba}
\end{figure}

\paragraph{Receptive-field intuition.}
We include a simplified linear analysis to clarify the spatial bias of WovenMamba for intra-image aggregation. In feature matching, contextual evidence should remain spatially structured: a local descriptor may need to distinguish whether supporting evidence appears above, to the left, or at a diagonal offset. The goal is not to model the full selective, gated, and input-dependent block, but to isolate the effect of composing a horizontal-order scan with a vertical-order scan.

Consider a one-dimensional linear state-space scan along an order $a\in\{h,v\}$:
\begin{equation}
    \mathbf{h}_a(t)
    =
    A_a\mathbf{h}_a(t-1)+B_a\mathbf{s}_a(t),
    \qquad
    \mathbf{o}_a(t)
    =
    C_a\mathbf{h}_a(t)+D_a\mathbf{s}_a(t).
\end{equation}
For nonnegative lags $\Delta\ge0$, this simplified scan induces the causal kernel
\begin{equation}
    K_a(\Delta)
    =
    C_a A_a^{\Delta} B_a
    +
    D_a\mathbf{1}[\Delta=0],
\end{equation}
with negative lags treated as zero. For an $H\times W$ grid, define
\begin{equation}
    \pi_h(i,j)=iW+j,
    \qquad
    \pi_v(i,j)=jH+i .
\end{equation}
The exact finite-grid kernel of the simplified horizontal-then-vertical composition is
\begin{equation}
\label{eq:woven_exact_kernel}
    W^{\mathrm{exact}}_{(i,j)\leftarrow(p,q)}
    =
    \sum_{r,s}
    K_v\!\left(\pi_v(i,j)-\pi_v(r,s)\right)
    K_h\!\left(\pi_h(r,s)-\pi_h(p,q)\right).
\end{equation}
Thus, information from $(p,q)$ reaches $(i,j)$ through intermediate grid positions $(r,s)$: the second scan acts on features already contextualized by the first scan.

\begin{figure}[t]
    \centering    \includegraphics[width=0.9\linewidth]{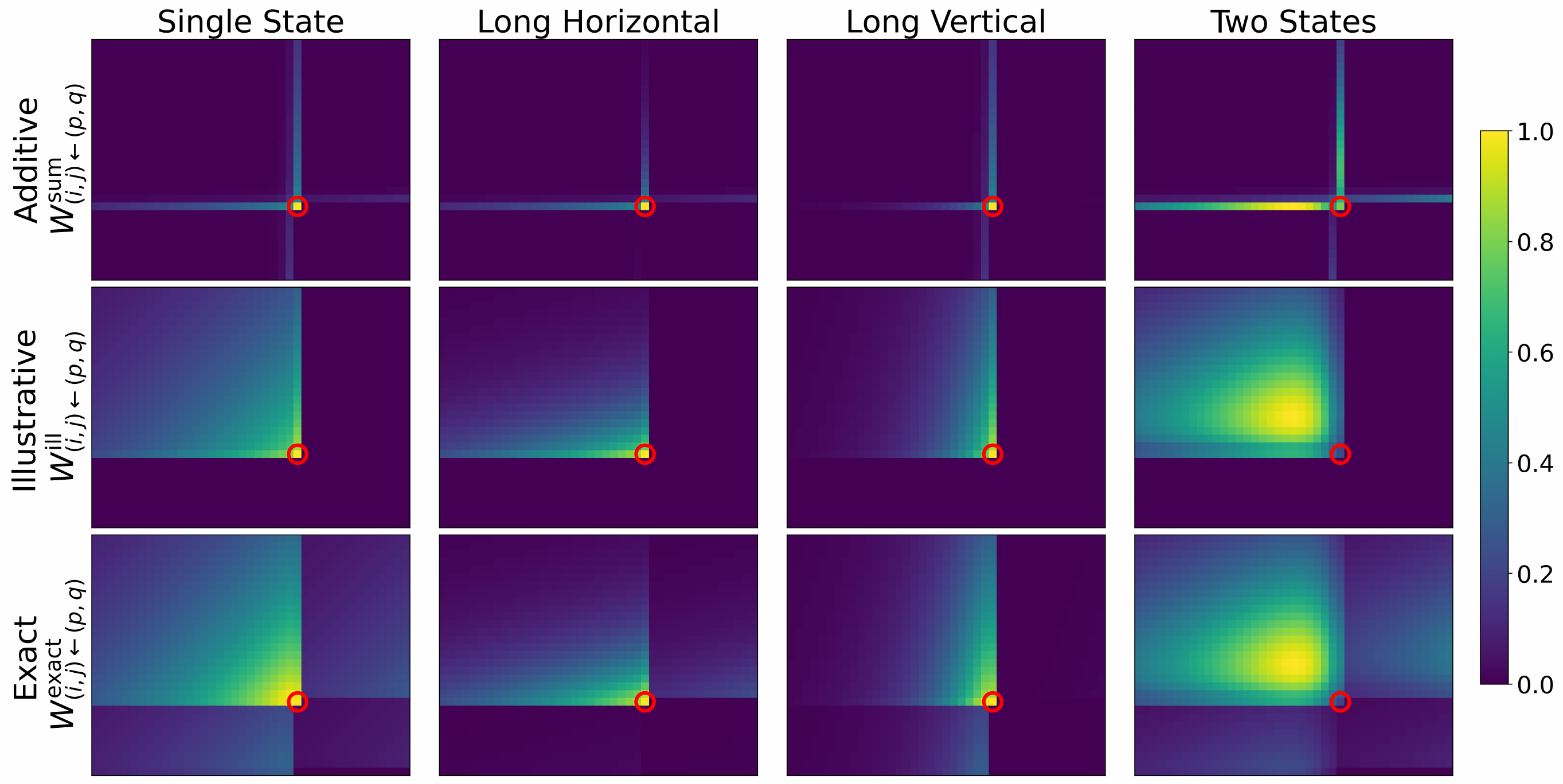}
    \caption{\textbf{Receptive-field intuition under simplified linear models.}
    The red circle marks the output position $(i,j)$, and color indicates the normalized influence from each source position $(p,q)$. Rows compare the additive scan-and-sum baseline $W^{\mathrm{sum}}$, the dominant-path approximation $W^{\mathrm{ill}}$, and the exact finite-grid kernel $W^{\mathrm{exact}}$ of the simplified WovenMamba composition. Columns show different one-dimensional decay settings: Single State, Long Horizontal, Long Vertical, and Two States.}
    \label{fig:receptive_field}
\end{figure}

For visualization, we also show a dominant-path approximation that keeps the intuitive same-row then same-column route:
\begin{equation}
\label{eq:woven_illustrative_kernel}
    W^{\mathrm{ill}}_{(i,j)\leftarrow(p,q)}
    =
    \mathbf{1}[p\le i]\mathbf{1}[q\le j]\,
    K_v(i-p)K_h(j-q).
\end{equation}
This approximation is not a separate module; it is only used to make the composition effect easier to interpret. The main comparison in Figure~\ref{fig:receptive_field} is between the additive baseline $W^{\mathrm{sum}}$ and the exact composed kernel $W^{\mathrm{exact}}$.

By contrast, a scan-and-sum design applies horizontal and vertical scans independently and adds their outputs:
\begin{equation}
    W^{\mathrm{sum}}_{(i,j)\leftarrow(p,q)}
    =
    K_h\!\left(\pi_h(i,j)-\pi_h(p,q)\right)
    +
    K_v\!\left(\pi_v(i,j)-\pi_v(p,q)\right).
\end{equation}
This additive form overlays independent one-dimensional responses, whereas WovenMamba composes them into direction- and distance-dependent two-dimensional support. Section~\ref{sec:visualizations} complements this simplified analysis with effective receptive fields from trained models.

\subsection{AnchorCrossAttention for inter-image aggregation}

\begin{figure}[t]
    \centering
    \includegraphics[width=0.98\linewidth]{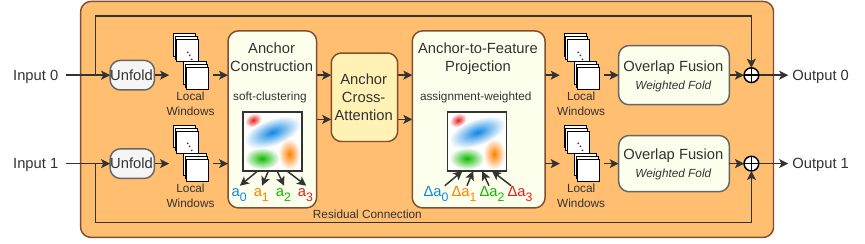}
    \caption{\textbf{AnchorCrossAttention.} Anchor construction pools local window features into anchors using soft clustering. After cross-image attention, anchor-to-feature projection distributes the anchor updates using the same assignment probabilities. Overlap fusion combines the window-specific updates on the dense grid through a weighted Fold using normalized overlap weights.}
    \label{fig:anchor_cross_attention}
\end{figure}

AnchorCrossAttention (ACA) complements WovenMamba's intra-image context aggregation by retrieving evidence from potentially corresponding regions in the paired image. Because individual token affinities can be ambiguous due to noise or viewpoint changes, ACA allows dense tokens to retrieve context through anchors that aggregate local feature evidence \citep{bezdek1981fuzzy,Arandjelovic2016NetVLAD,lee2019set}. As shown in Figure~\ref{fig:anchor_cross_attention}, ACA performs anchor construction, Anchor CrossAttention, and anchor-to-feature projection, followed by overlap fusion on the dense feature grid.

\paragraph{Anchor construction.}
A local window can contain different structures, so we use multiple anchors to support separate contextual updates. We construct anchors by soft clustering within each window, using learned assignment probabilities to pool the features. Let $\widetilde{\mathbf{F}}_c^0,\widetilde{\mathbf{F}}_c^1\in\mathbb{R}^{H_c\times W_c\times C_c}$ denote the coarse feature maps after WovenMamba. We extract $M$ overlapping windows of size $w\times w$ with stride $w/2$. Let $\mathcal{W}_m$ denote the set of coarse-grid positions in the $m$-th window, where $m\in\{1,\ldots,M\}$. For each position $(u,v)\in\mathcal{W}_m$, we compute the soft assignment probability $P_{m,k}(u,v)$ of its feature to anchor $k\in\{0,\ldots,K-1\}$. These assignments are computed separately for each image; we omit the image index for clarity:
\begin{equation}
    P_{m,k}(u,v)
    =
    \operatorname{softmax}_{k}
    \left(
    \mathbf{q}_k^\top
    \bar{\mathbf{x}}_{m}(u,v)/\tau
    \right),
\end{equation}
where $\bar{\mathbf{x}}_{m}(u,v)\in\mathbb{R}^{C_c}$ denotes the window-wise normalized feature at $(u,v)$ and $\tau>0$ is a learned temperature. The clustering vectors $\mathbf{q}_k\in\mathbb{R}^{C_c}$ are shared across all windows and both images within each ACA block, providing a common assignment parameterization. Each window produces an assignment matrix of size $w^2\times K$. For anchor $k$, we pool raw and normalized features using $P_{m,k}(u,v)$ as weights, normalizing by the total assignment weight, to obtain $\mathbf{f}^{\mathrm{raw}}_{m,k},\,\mathbf{f}^{\mathrm{norm}}_{m,k}\in\mathbb{R}^{C_c}$. Because pooling discards spatial arrangement, we also encode the assignment center, covariance, and mass into $\mathbf{f}^{\mathrm{geo}}_{m,k}\in\mathbb{R}^{C_g}$. The anchor descriptor is
\begin{equation}
    \mathbf{a}_{m,k}
    =
    \operatorname{Linear}_{\mathrm{anc}}
    \left(
    [\mathbf{f}^{\mathrm{raw}}_{m,k},
    \mathbf{f}^{\mathrm{norm}}_{m,k},
    \mathbf{f}^{\mathrm{geo}}_{m,k}]
    \right)
    \in\mathbb{R}^{D}.
\end{equation}
Appendix~\ref{app:aca_details} provides the detailed pooling and geometry formulas.

\paragraph{Anchor CrossAttention.}
For each anchor ID $k$, we stack the descriptors $\mathbf{a}^i_{m,k}$ from the $M$ windows as the rows of $\mathbf{A}^i_k\in\mathbb{R}^{M\times D}$, where $i\in\{0,1\}$ is the image index. Anchor-level attention connects each anchor to all $M$ anchors with the same ID in the other image. We compute
\begin{equation}
    \mathbf{C}^i_k
    =
    \operatorname{Attn}
    (\mathbf{A}^i_k,\mathbf{A}^{1-i}_k,\mathbf{A}^{1-i}_k),
    \qquad
    \Delta\mathbf{A}^i_k
    =
    \operatorname{Linear}_{\Delta}
    (\mathbf{A}^i_k+\mathbf{C}^i_k),
\end{equation}
where $\operatorname{Attn}(\mathbf{Q},\mathbf{K},\mathbf{V})$ denotes multi-head cross-attention \citep{vaswani2017attention} with query, key, and value inputs $\mathbf{Q}$, $\mathbf{K}$, and $\mathbf{V}$, respectively, and $\Delta\mathbf{A}^i_k\in\mathbb{R}^{M\times C_c}$ contains the anchor feature updates. Its $m$-th row, $\Delta\mathbf{a}^i_{m,k}$, is the update for anchor $k$ in window $m$. Grouping by ID reduces anchor-attention interactions from $O(K^2M^2)$ for joint attention over all anchors to $O(KM^2)$, while retaining access to all windows in the paired image.

\paragraph{Anchor-to-feature projection and overlap fusion.}
Anchor-to-feature projection combines the updates from the $K$ anchors at each position within a window using the same assignment probabilities. Since the concentration of a token's anchor assignments can vary across overlapping windows, we use assignment entropy as a cue to learn the relative contributions of the window-specific updates. Let $\mathcal{I}(u,v)=\{m\mid (u,v)\in\mathcal{W}_m\}$ denote the indices of windows containing $(u,v)$ and $\lambda^i_m(u,v)$ the overlap weights normalized over those windows. The updated coarse features are
\begin{equation}
\label{eq:anchor_to_dense}
    \widehat{\mathbf{F}}_c^i(u,v)
    =
    \widetilde{\mathbf{F}}_c^i(u,v)
    +
    \sum_{m\in\mathcal{I}(u,v)}
    \lambda^i_{m}(u,v)
    \sum_{k=0}^{K-1}
    P^i_{m,k}(u,v)
    \Delta\mathbf{a}^i_{m,k}.
\end{equation}
The inner sum performs assignment-weighted anchor-to-feature projection, while the outer weighted sum implements overlap fusion as a weighted Fold. Appendix~\ref{app:aca_details} provides the overlap-weighting formulas, component ablations, and assignment analysis.

\section{Experiments}
\label{sec:experiments}

\subsection{Experimental setup}
\label{sec:experimental_setup}

\paragraph{Implementation.}
WAM follows the overall framework family of JamMa \citep{lu2025jamma} with a lightweight ConvNeXtV2-style encoder \citep{woo2023convnextv2}. Coarse features at $1/8$ resolution are updated by two WovenAnchorModule stages, while fine features at $1/2$ resolution are used for local refinement. \aca uses $8\times8$ windows with a stride of 4 and $K=4$ anchors per window. We train on MegaDepth images resized and padded to $832\times832$ \citep{li18megadepth} for 20 epochs with batch size 8, using AdamW with a base learning rate of $4\times10^{-4}$, linear warm-up for 14,550 optimization steps (approximately 3.16 epochs), and cosine annealing. Training takes about 26 hours on a single NVIDIA H100 GPU.
Appendix~\ref{app:implementation} provides the remaining hyperparameters, matching thresholds, RANSAC settings, and runtime protocol.


\paragraph{Evaluation.}
We evaluate relative pose estimation on MegaDepth-1500, an outdoor benchmark based on MegaDepth \citep{li18megadepth}, and zero-shot relative pose estimation on ScanNet-1500, an indoor benchmark based on ScanNet \citep{dai2017scannet}, following the LoFTR/JamMa protocol \citep{sun21loftr,lu2025jamma}. MegaDepth images are resized and padded to $832\times832$; ScanNet images are resized to $640\times480$. We use LO-RANSAC \citep{Chum2003LORANSAC} and report pose AUC at $5^\circ$, $10^\circ$, and $20^\circ$. For zero-shot homography evaluation on HPatches, a planar homography benchmark with viewpoint and illumination changes \citep{balntas17hpatches}, we follow LoFTR/JamMa: resize to a 480-pixel short edge, retain the top 1000 matches by confidence, and use vanilla RANSAC \citep{fischler1981RANSAC}. We report corner-error AUC at 3, 5, and 10 pixels. Runtimes are measured with batch size 1 on the same H100 GPU, including feature extraction and matching but excluding image I/O and RANSAC.

\subsection{Benchmark comparison}
\label{sec:benchmark_comparison}

\begin{table}[t]
\centering
\caption{\textbf{MegaDepth-1500 relative pose estimation.}
We report parameters, runtime, and pose-error AUC. Best and second-best semi-dense results are highlighted. 
XFeat$^{*}$ denotes the semi-dense variant with match refinement.
WAM$^\dagger$ reports the checkpoint whose AUC@5$^\circ$ equals the median over seven runs; full repeated-run statistics are in Appendix~\ref{app:repeated_runs}.}
\label{tab:main_megadepth}
\begin{tabular}{clcc*{3}{>{\centering\arraybackslash}p{30pt}}}
\toprule
\multirow{2}{*}{Model family} & \multirow{2}{*}{Matcher} & Params & Runtime & \multicolumn{3}{c}{\parbox[c]{\dimexpr90pt+4\tabcolsep\relax}{\centering Relative pose error AUC $\uparrow$}} \\
\cmidrule(lr){5-7}
 & & (M) & (ms) & @5$^\circ$ & @10$^\circ$ & @20$^\circ$ \\
\midrule
\multirow{3}{*}{Sparse}
& XFeat \citep{potje24xfeat} & 0.7 & 12.5 & 44.2 & 58.2 & 69.2 \\
& SP+SuperGlue \citep{sarlin20superglue} & 13.3 & 55.2 & 57.6 & 72.6 & 83.5 \\
& SP+LightGlue \citep{sarlin23lightglue} & 13.2 & 47.2 & 58.8 & 73.6 & 84.1 \\
\midrule
\multirow{2}{*}{Dense}
& DKM \citep{edstedt23dkm} & 72.3 & 315.9 & 67.3 & 79.7 & 88.1 \\
& RoMa \citep{edstedt24roma} & 111.3 & 470.2 & 68.5 & 80.6 & 88.8 \\
\midrule
\multirow{9}{*}{Semi-dense}
& XFeat$^{*}$ \citep{potje24xfeat} & \textbf{1.5} & \underline{25.4} & 50.8 & 66.8 & 78.8 \\
& RCM \citep{lu24rcm} & 9.8 & 53.0 & 58.3 & 72.8 & 83.5 \\
& LoFTR \citep{sun21loftr} & 11.6 & 63.6 & 62.1 & 75.6 & 85.0 \\
& MatchFormer \citep{wang22matchformer} & 20.3 & 106.0 & 62.0 & 75.6 & 84.9 \\
& ASpanFormer \citep{chen22aspanformer} & 15.8 & 91.9 & 62.6 & 76.1 & 85.7 \\
& ELoFTR \citep{wang24eloftr} & 16.0 & 41.5 & 63.7 & 77.0 & 86.4 \\
& JamMa \citep{lu2025jamma} & 5.7 & 43.8 & 64.1 & 77.4 & 86.5 \\
& EDM \citep{Li2025EDM} & 10.3 & \textbf{20.9} & \underline{64.5} & \underline{77.6} & \underline{86.8} \\
& WAM (Ours)$^\dagger$ & \underline{4.5} & 35.4 & \textbf{66.1} & \textbf{78.9} & \textbf{87.7} \\
\bottomrule
\end{tabular}
\end{table}


Table~\ref{tab:main_megadepth} reports MegaDepth relative pose estimation. WAM follows a JamMa-style coarse-to-fine pipeline, with its architectural contribution centered on role-specialized intra-image and inter-image context aggregation. The WAM$^\dagger$ row reports the checkpoint whose AUC@5$^\circ$ equals the median over seven runs; it achieves 66.1 AUC@5$^\circ$, compared with 64.1 for JamMa and 64.5 for EDM. WAM's full-matcher runtime is 35.4 ms, compared with 43.8 ms for JamMa. This runtime difference primarily comes from coarse-matching optimizations, while WovenAnchorModule itself is more expensive than JointMamba. Appendix~\ref{app:implementation} provides the detailed runtime breakdown and implementation explanation. Figure~\ref{fig:tradeoff} summarizes the accuracy--runtime comparison.

Appendix~\ref{app:spatial_resolutions} details the spatial representations used by WovenAnchorModule and coarse matching.

Table~\ref{tab:zero_shot_transfer} reports zero-shot relative pose estimation on ScanNet and homography estimation on HPatches using MegaDepth-trained weights, without fine-tuning on either benchmark. WAM improves over JamMa on both benchmarks and remains competitive with EDM, suggesting that the proposed aggregation transfers beyond MegaDepth.

\begin{table*}[t]
\centering
\caption{\textbf{Zero-shot transfer using MegaDepth-trained weights.}
ScanNet reports relative pose AUC at angular error thresholds; HPatches reports homography AUC at corner-error thresholds in pixels.}
\label{tab:zero_shot_transfer}
\setlength{\tabcolsep}{3.5pt}
\begin{tabular}{@{}l*{6}{>{\centering\arraybackslash}p{45pt}}@{}}
\toprule
\multirow{2}{*}{Matcher} & \multicolumn{3}{c}{\parbox[c]{\dimexpr135pt+4\tabcolsep\relax}{\centering ScanNet relative pose error AUC~$\uparrow$}} & \multicolumn{3}{c@{}}{\parbox[c]{\dimexpr135pt+4\tabcolsep\relax}{\centering HPatches homography error AUC~$\uparrow$}} \\
\cmidrule(lr){2-4}
\cmidrule(lr){5-7}
 & @5$^\circ$ & @10$^\circ$ & @20$^\circ$ & @3px & @5px & @10px \\
\midrule
SP+SuperGlue \citep{sarlin20superglue} & 16.5 & 32.2 & 48.8 & 53.9 & 68.3 & 81.7 \\
SP+LightGlue \citep{sarlin23lightglue} & 15.9 & 32.2 & 49.7 & 54.2 & 68.3 & 81.5 \\
LoFTR \citep{sun21loftr} & 20.7 & 37.7 & 53.4 & 65.9 & 75.6 & 84.6 \\
ASpanFormer \citep{chen22aspanformer} & 21.9 & 40.6 & \textbf{57.4} & 67.4 & 76.9 & 85.6 \\
JamMa \citep{lu2025jamma} & 18.5 & 35.6 & 51.3 & \underline{68.1} & 77.0 & 85.4 \\
EDM \citep{Li2025EDM} & \underline{22.7} & \underline{40.8} & 56.9 & \textbf{68.5} & \textbf{78.1} & \textbf{86.6} \\
WAM (Ours) & \textbf{23.2} & \textbf{41.2} & \underline{57.2} & \textbf{68.5} & \underline{77.9} & \underline{86.4} \\
\bottomrule
\end{tabular}
\end{table*}

\subsection{Controlled ablations}
\label{sec:ablations}

The ablations are conducted under the same setup. The component ablation first shows that both intra-image and inter-image context aggregation are useful. The alternative-module comparisons and the shared/reversed-role experiment then evaluate which operators are better suited to each role.

\paragraph{Component ablation.}
Table~\ref{tab:woven_anchor_module} removes the two context aggregation components of WAM. Removing WovenMamba reduces AUC@5$^\circ$ from 66.1 to 64.6, while removing AnchorCrossAttention reduces it to 64.0. Removing both gives 62.0. These results show that both intra-image and inter-image context aggregation are important.

\begin{table}[t]
\centering
\caption{\textbf{Component ablations on MegaDepth-1500.} We report pose-error AUC under WAM component removals. Runtime is measured for the full matcher.}
\label{tab:woven_anchor_module}
\setlength{\tabcolsep}{3.5pt}
\begin{tabular}{@{}lc*{3}{>{\centering\arraybackslash}p{40pt}}@{}}
\toprule
\multirow{2}{*}{Variant}
& Runtime
& \multicolumn{3}{c@{}}{
    \parbox[c]{\dimexpr120pt+4\tabcolsep\relax}{
        \centering Relative pose error AUC $\uparrow$
    }
} \\
\cmidrule(lr){3-5}
 & (ms) & @5$^\circ$ & @10$^\circ$ & @20$^\circ$ \\
\midrule
WovenAnchorModule (Default) & 35.4 & \textbf{66.1} & \textbf{78.9} & \textbf{87.7} \\
w/o WovenMamba & 30.5 & 64.6 & 77.7 & 86.6 \\
w/o AnchorCrossAttention & 27.4 & 64.0 & 77.0 & 86.1 \\
w/o both & 20.9 & 62.0 & 75.0 & 84.7 \\
\bottomrule
\end{tabular}
\end{table}

\paragraph{Intra-image alternatives.}
Table~\ref{tab:abl_intra_image} compares intra-image context modules while keeping AnchorCrossAttention fixed. UniMamba, BiMamba, OrthMamba, and FourMamba are directional-scan baselines with one, two opposite, two orthogonal, and four independent scan directions, respectively. All scan variants are below WovenMamba, and FourMamba increases runtime without matching WovenMamba's accuracy. Self-Attention serves as an attention-based intra-image alternative, but it is slower and less accurate. These results suggest that the proposed woven horizontal-then-vertical propagation gives a better accuracy--efficiency trade-off than either independent directional scans or self-attention.

\paragraph{Inter-image alternatives.}
Table~\ref{tab:abl_inter_image} compares inter-image modules with WovenMamba fixed. The CrossMamba variants process the paired image with a Mamba scan and inject the resulting features through a residual update. They show that context from the paired image can help to some extent, but they lack explicit correspondence-aware retrieval based on cross-image feature affinities. Dense CrossAttention yields lower AUC@5$^\circ$ and higher full-matcher runtime than ACA under the same setup. AnchorCrossAttention achieves the best accuracy among these alternatives, supporting locally grouped anchors for inter-image aggregation.

\begin{table*}[t]
\centering
\setlength{\tabcolsep}{4.5pt}
\renewcommand{\arraystretch}{0.98}

\begin{minipage}[t]{0.43\linewidth}
\centering
\captionof{table}{\textbf{Intra-image ablations on MegaDepth-1500.} ACA is fixed. Runtime is measured for the full matcher.}
\label{tab:abl_intra_image}
\begin{tabular}{lcccc}
\toprule
\multirow{2}{*}{Variant} & Time & \multicolumn{3}{c}{Pose AUC $\uparrow$} \\
\cmidrule(lr){3-5}
 & (ms) & 5$^\circ$ & 10$^\circ$ & 20$^\circ$ \\
\midrule
WovenMamba & 35.4 & \textbf{66.1} & \textbf{78.9} & \textbf{87.7} \\
UniMamba & 33.1 & 65.2 & 78.3 & 87.3 \\
BiMamba & 35.7 & 65.2 & 78.4 & 87.3 \\
OrthMamba & 36.0 & 65.2 & 78.2 & 87.3 \\
FourMamba & 40.4 & 65.3 & 78.5 & 87.4 \\
Self-Attention & 49.0 & 64.8 & 77.8 & 86.9 \\
\bottomrule
\end{tabular}
\end{minipage}
\hfill
\begin{minipage}[t]{0.53\linewidth}
\centering
\captionof{table}{\textbf{Inter-image ablations on MegaDepth-1500.} WovenMamba is fixed. Runtime is measured for the full matcher.}
\label{tab:abl_inter_image}
\begin{tabular}{lcccc}
\toprule
\multirow{2}{*}{Variant} & Time & \multicolumn{3}{c}{Pose AUC $\uparrow$} \\
\cmidrule(lr){3-5}
 & (ms) & 5$^\circ$ & 10$^\circ$ & 20$^\circ$ \\
\midrule
AnchorCrossAttention & 35.4 & \textbf{66.1} & \textbf{78.9} & \textbf{87.7} \\
CrossUniMamba & 29.7 & 65.3 & 78.3 & 87.1 \\
CrossBiMamba & 34.6 & 65.3 & 78.2 & 87.1 \\
CrossOrthMamba & 32.3 & 65.3 & 78.2 & 87.0 \\
CrossFourMamba & 37.1 & 65.4 & 78.2 & 87.2 \\
Dense CrossAttention & 50.3 & 65.0 & 78.2 & 87.4 \\
\bottomrule
\end{tabular}
\end{minipage}

\end{table*}

\paragraph{Operator-to-role assignment.}
Table~\ref{tab:role_assignment} tests whether the gains depend on the operator-to-role assignment. AnchorSelfAttention (ASA) applies the anchor operator within each image, while CrossWovenMamba (CWM) adapts WovenMamba to exchange context between images. Using the same operator family for both roles or reversing their assignment reduces pose AUC relative to WovenMamba for intra-image propagation and ACA for inter-image retrieval. These results support the proposed specialization of the two operators under the evaluated setup. Appendix~\ref{app:implementation} describes the adaptations.

\begin{table}[t]
\centering
\caption{\textbf{Operator-to-role assignment on MegaDepth-1500.} ASA: AnchorSelfAttention; CWM: CrossWovenMamba. Runtime is measured for the full matcher.}
\label{tab:role_assignment}
\setlength{\tabcolsep}{3.5pt}
\begin{tabular}{llcccc}
\toprule
Intra-image & Inter-image & AUC@5$^\circ$ & AUC@10$^\circ$ & AUC@20$^\circ$ & Runtime (ms) \\
\midrule
WovenMamba & ACA (default) & \textbf{66.1} & \textbf{78.9} & \textbf{87.7} & 35.4 \\
ASA & ACA & 64.7 & 77.8 & 86.7 & 38.6 \\
WovenMamba & CWM & 65.4 & 78.3 & 87.4 & 33.9 \\
ASA & CWM & 64.9 & 77.7 & 86.7 & 36.0 \\
\bottomrule
\end{tabular}
\end{table}

\subsection{Visualizations}
\label{sec:visualizations}

\begin{figure*}[t]
    \centering
    \includegraphics[width=\linewidth]{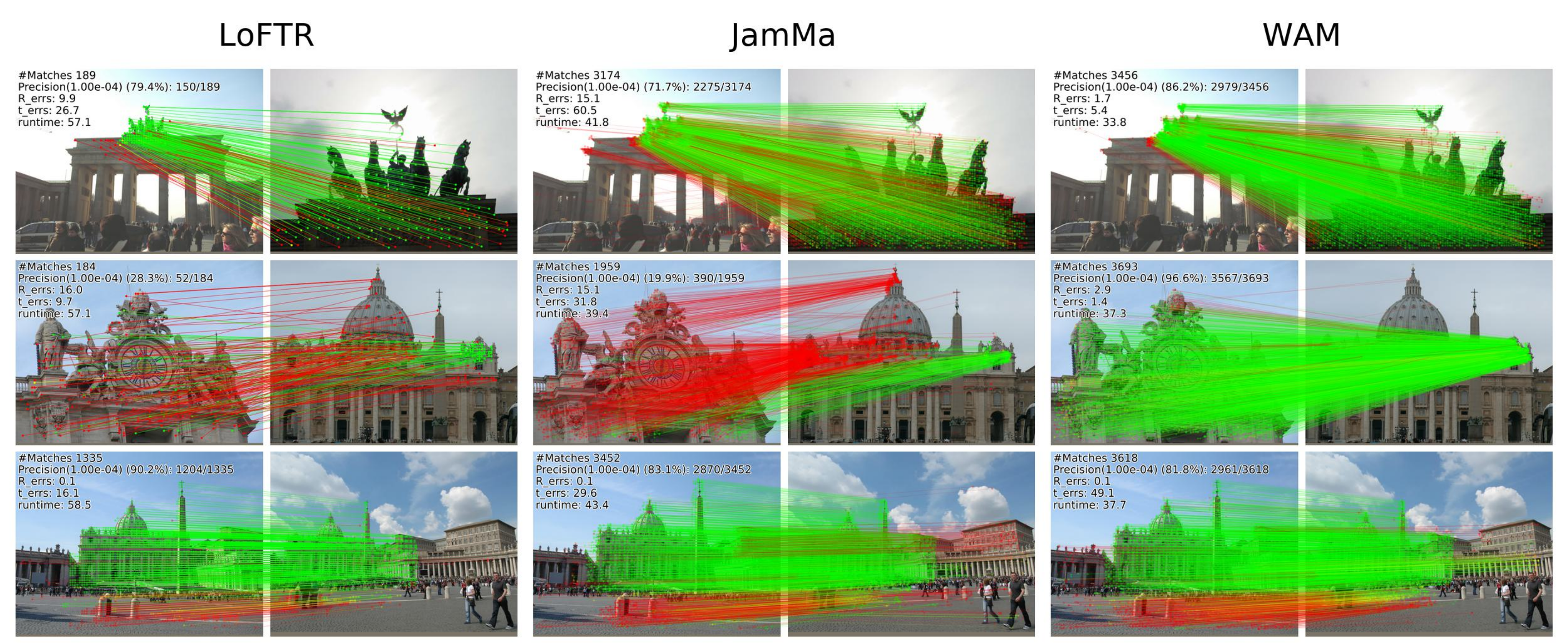}
    \caption{\textbf{Qualitative matching examples.} We compare LoFTR, JamMa, and WAM. Green and red lines indicate geometrically consistent and inconsistent matches, respectively.}
    \label{fig:qualitative}
\end{figure*}

\paragraph{Qualitative matching results.}
Figure~\ref{fig:qualitative} shows representative successes and a failure case. In the first two pairs, WAM produces dense and geometrically consistent matches under large appearance or viewpoint changes. The last pair shows a limitation: repetitive or weakly constrained regions can still lead to incorrect matches.

\paragraph{Learned anchor assignments.}
Appendix~\ref{app:aca_visualizations} visualizes the learned soft assignments used to construct anchors for two image pairs. Brighter heatmap colors indicate higher assignment probabilities to the corresponding anchor; categorical colors identify the highest-probability anchor. In these examples, the first ACA block (ACA-1) shows comparatively diffuse responses, although different anchors already respond to low-texture regions and contours. The second ACA block (ACA-2) exhibits more distinct spatial preferences: anchor 1 often emphasizes the structure side of building and statue boundaries, while anchor 3 emphasizes adjacent exterior regions, particularly the sky side. Anchors 0 and 2 show broader responses over low-texture surfaces and less structured regions. These spatial preferences suggest that the anchors summarize different local structures within each window. Projecting their cross-image updates through the learned assignments allows tokens associated with different structures to receive different contextual information. This may help disambiguate visually similar features during matching. The larger assignment variances in ACA-2 (Table~\ref{tab:anchor_assignment_stats}) provide complementary evidence of greater variation in assignment strength.

\paragraph{Learned effective receptive fields.}
Appendix~\ref{app:woven_erf} compares local effective receptive fields (ERFs) of trained horizontal-then-vertical WovenMamba and the OrthMamba scan-and-sum baseline. Brighter colors indicate greater relative sensitivity of the target's residual update to the block input, normalized separately for each target and shown on a logarithmic scale. In the examples shown, WovenMamba emphasizes a coherent two-dimensional neighborhood above and to the left of the target, while OrthMamba exhibits more pronounced ridges along the target's row and column together with diffuse, content-dependent responses away from these axes. The WovenMamba pattern is qualitatively consistent with the directional and distance-dependent bias suggested by the simplified composition analysis. Such spatially structured context may help distinguish locations with similar local appearances by incorporating the target's neighboring structure. These observations complement the controlled intra-image ablations in Table~\ref{tab:abl_intra_image}.

\section{Discussion and limitations}
\label{sec:discussion_and_limitations}
The additional visual localization evaluation in Appendix~\ref{app:aachen} provides a diagnostic view of how WAM behaves in a larger localization pipeline. The experiments support the value of specializing intra-image and inter-image aggregation: the simplified receptive-field analysis, trained-model ERFs, and intra-image ablations support horizontal-then-vertical propagation, while the inter-image ablations support local anchor grouping for correspondence-aware retrieval. A limitation is computational cost and scalability. WovenAnchorModule has a higher measured runtime than JointMamba in our evaluation (Table~\ref{tab:runtime_breakdown}). Although AnchorCrossAttention reduces the number of interactions relative to dense token-wise cross-attention, its interaction complexity remains quadratic in the number of local windows, limiting scalability at high input resolutions. Another limitation is the JamMa-style nearest-neighbor matching strategy, which may produce many-to-one correspondences and may contribute to the localization gap between WAM and LoFTR in the Aachen pipeline. WAM generally improves over JamMa under the same matching strategy, but these results do not isolate the effect of match filtering.

\section{Conclusion}

We presented WAM, a semi-dense matcher that specializes context aggregation for intra-image and inter-image modeling. WovenMamba organizes spatial context with horizontal-then-vertical SSM propagation, while AnchorCrossAttention retrieves cross-image context through locally grouped anchors. A simplified receptive-field analysis suggests that WovenMamba provides a structural bias toward composed two-dimensional dependencies. Controlled ablations show that both forms of context are useful and that the proposed specialized designs are better suited to their respective roles. Benchmark results further show that WAM improves over JamMa under our setup. Overall, these results support designing semi-dense matchers with context modules matched to the different requirements of intra-image and inter-image aggregation.

\newpage
\begin{ack}
This work was supported by DENSO IT LAB Recognition, Control, and Learning Algorithm Collaborative Research Chair (Science Tokyo) and the experiments were conducted using TSUBAME 4.0 supercomputer.
The authors declare no competing interests.
\end{ack}

\bibliographystyle{unsrtnat}
\bibliography{references}

\newpage
\appendix
\section{Implementation and evaluation details}
\label{app:implementation}

\paragraph{Architecture details.}
Unless otherwise stated, WAM uses a coarse feature dimension of $C_c=256$ and a fine feature dimension of 64. The WovenMamba propagation branch has dimension 512, corresponding to an expansion ratio of 2. WovenMamba uses state dimension 8 and $3\times3$ depthwise convolutions before the horizontal and vertical SSM scans. RMSNorm is used in all WovenMamba and AnchorCrossAttention blocks. AnchorCrossAttention uses overlapping windows of size $8\times8$ with a stride of 4, $K=4$ anchors per window, anchor dimension $D=256$, geometry feature dimension $C_g=16$, 4 attention heads, and zero attention dropout. We use two WovenAnchorModule stages unless otherwise stated.

\paragraph{WovenMamba separator tokens.}
For the reported WovenMamba models, we insert fixed separator tokens into the flattened scan sequences at scan-boundary positions: row boundaries for the horizontal-order scan and column boundaries for the vertical-order scan. The SSM is applied to the augmented sequence; after unscanning the augmented output, we remove the separator positions and keep only the original image-grid tokens. At separator positions, we add a fixed offset of $6.0$ to the pre-softplus SSM step-size parameter. This is a boundary-handling implementation detail and is not treated as a separate contribution.

\begin{table}[h]
\centering
\caption{\textbf{Effect of WovenMamba separator tokens.}}
\label{tab:separator_tokens}
\setlength{\tabcolsep}{3.2pt}
\begin{tabular}{lcccccc}
\toprule
\multirow{2}{*}{Variant} &
\multicolumn{3}{c}{MegaDepth-1500} &
\multicolumn{3}{c}{ScanNet-1500} \\
\cmidrule(lr){2-4}
\cmidrule(lr){5-7}
& AUC@5$^\circ$ & AUC@10$^\circ$ & AUC@20$^\circ$ &
AUC@5$^\circ$ & AUC@10$^\circ$ & AUC@20$^\circ$ \\
\midrule
w/o separators & 66.0 & 78.8 & 87.7 & 22.5 & 40.6 & 56.5 \\
with separators & 66.1 & 78.9 & 87.7 & 23.2 & 41.2 & 57.2 \\
\bottomrule
\end{tabular}
\end{table}
Table~\ref{tab:separator_tokens} shows that the separators have little effect on MegaDepth accuracy, but improve the zero-shot ScanNet results in our evaluation. We therefore use them in the reported model as a small robustness-oriented implementation detail, while keeping the main claims focused on WovenMamba and AnchorCrossAttention.

\paragraph{Matching details.}
For coarse matching, we apply softmax independently along the rows and columns of the similarity matrix with a temperature of 0.1. We retain directional nearest-neighbor matches with probability greater than 0.2 and take the union of the two sets, assigning the larger directional probability to duplicate pairs. For fine matching, we use a fine matching window of size $5\times5$, dual-softmax temperature 0.1, and confidence threshold 0.1. Unless otherwise stated, the same matching thresholds are used for the main benchmark results and ablations.

\paragraph{Training details.}
The main model is trained on MegaDepth for 20 epochs with batch size 8 using the training protocol described in Section~\ref{sec:experimental_setup}. Training uses a single NVIDIA H100 GPU; the peak allocated GPU memory for the main model is about 85GB. Each WAM training run takes about 26 hours on one H100 GPU. The sub-pixel refinement loss uses a squared symmetric epipolar distance threshold of $10^{-4}$ on MegaDepth. The main benchmark tables use a representative WAM checkpoint whose MegaDepth AUC@5$^\circ$ equals the median over seven runs, as reported in Appendix~\ref{app:repeated_runs}. Results labeled as zero-shot transfer evaluate this representative MegaDepth-trained checkpoint on the target dataset without additional fine-tuning.

\paragraph{Ablation protocol.}
All ablation variants are trained on MegaDepth using the same training protocol as the main model, unless otherwise stated. 
We change only the component specified in each ablation table and keep the backbone, feature resolutions, number of WovenAnchorModule stages, optimizer, learning-rate schedule, batch size, matching thresholds, and evaluation protocol fixed. 
This ensures that the ablations isolate the contribution of the modified component rather than changes in the training or evaluation setup. 
Ablation variants are single-run controlled trainings unless repeated-run statistics are explicitly reported.

\paragraph{Role-assignment adaptations.}
For Table~\ref{tab:role_assignment}, AnchorSelfAttention retains ACA's anchor construction, anchor-to-feature projection, and overlap fusion, but replaces cross-image attention with self-attention over the windows of each image within each anchor-ID group. CrossWovenMamba computes WovenMamba's residual updates separately for the two images and adds each update to the paired image's features. The shared and reversed variants retain two stages and the same feature dimensions and training/evaluation protocol as the default model.

\paragraph{Pose and homography evaluation.}
For relative pose estimation on MegaDepth and ScanNet, we estimate the essential matrix using LO-RANSAC \citep{Chum2003LORANSAC} implemented in PoseLib 2.0.4 with a threshold of 0.5 pixels, leaving the remaining RANSAC options at their library defaults. Pose accuracy is reported using the area under the cumulative error curve at $5^\circ$, $10^\circ$, and $20^\circ$, following the standard protocol for these benchmarks. For homography estimation on HPatches, we retain the top 1000 matches by confidence and estimate the homography using vanilla RANSAC \citep{fischler1981RANSAC} with threshold 3.0, confidence 0.99999, and a maximum of 3000 iterations.

\paragraph{Runtime measurement.}
Unless otherwise stated, we measure runtime on a single NVIDIA H100 GPU with batch size 1 using MegaDepth image pairs resized and padded to $832\times832$. We perform 50 warm-up steps before timing and exclude them from the reported mean runtimes over the evaluated image pairs. End-to-end timing includes backbone feature extraction and matching, but excludes image I/O and RANSAC.

\paragraph{Runtime breakdown.}
Table~\ref{tab:runtime_breakdown} reports per-stage runtimes. WovenAnchorModule takes 14.0 ms, compared with 2.9 ms for JointMamba. WAM's lower end-to-end runtime primarily comes from implementation optimizations in coarse matching, which takes 4.7 ms for WAM and 20.9 ms for JamMa.
Specifically, coarse matching computes directional softmax maxima and their indices directly, without materializing dense probability matrices. This implementation choice contributes to full-matcher efficiency; the architectural contribution is role-specialized context aggregation.

\begin{table}[t]
\centering
\caption{\textbf{Runtime breakdown on MegaDepth (ms).} Stage timings exclude some overhead and therefore do not sum exactly to the end-to-end totals.}
\label{tab:runtime_breakdown}
\begin{tabular}{lcc}
\toprule
Component & WAM (ms) & JamMa (ms) \\
\midrule
Backbone & 5.6 & 5.9 \\
Context module & 14.0 & 2.9 \\
Coarse matching & 4.7 & 20.9 \\
Fine matching & 9.7 & 10.2 \\
\midrule
End-to-end total & 35.4 & 43.8 \\
\bottomrule
\end{tabular}
\end{table}

\section{Spatial resolutions in WAM}
\label{app:spatial_resolutions}

Table~\ref{tab:spatial_resolutions} summarizes the representations used by WAM. The backbone produces dense coarse features at $1/8$ of the input resolution. Both WovenAnchorModule stages preserve this grid: WovenMamba propagates context over its positions, and ACA constructs anchors from overlapping windows, then uses anchor-to-feature projection and overlap fusion to return their updates to the same grid. Coarse matching uses the resulting dense $1/8$ features.

With the default $8\times8$ windows and stride 4, adjacent ACA window centers are separated by four coarse cells, or 32 input pixels. Each window supplies $K=4$ anchors. Thus, the $1/32$ window-center spacing describes the sampling of anchor windows, rather than a dense $1/32$ feature map. For a padded $832\times832$ input, the coarse grid is $104\times104$ and ACA constructs $25\times25$ windows, giving 2500 anchors per image. Cross-attention is applied separately to the 625 anchors in each anchor-ID group. Padded image regions are masked; ACA's window-boundary padding and masking are described in Appendix~\ref{app:aca_details}.

\begin{table}[htbp]
\centering
\caption{\textbf{Spatial representations in WAM.} Grid sizes illustrate a padded $832\times832$ input with the default window and anchor settings.}
\label{tab:spatial_resolutions}
\setlength{\tabcolsep}{3.5pt}
\begin{tabular}{lll}
\toprule
Stage & Representation & Spatial sampling \\
\midrule
Before WovenAnchorModule & $104\times104$ coarse features & Dense $1/8$ grid \\
WovenMamba & $104\times104$ coarse features & Dense $1/8$ grid \\
ACA anchor construction & $8\times8$ coarse cells/window & Dense $1/8$ grid \\
ACA cross-attention & $25\times25$ windows, 4 anchors each & $1/32$ center spacing \\
\shortstack[l]{ACA anchor-to-feature projection\\and overlap fusion} & $104\times104$ coarse features & Dense $1/8$ grid \\
Coarse matching & $104\times104$ coarse features & Dense $1/8$ grid \\
\bottomrule
\end{tabular}
\end{table}

\section{Trained-model effective receptive fields}
\label{app:woven_erf}

\begin{figure}[htbp]
    \centering
    \includegraphics[width=\linewidth]{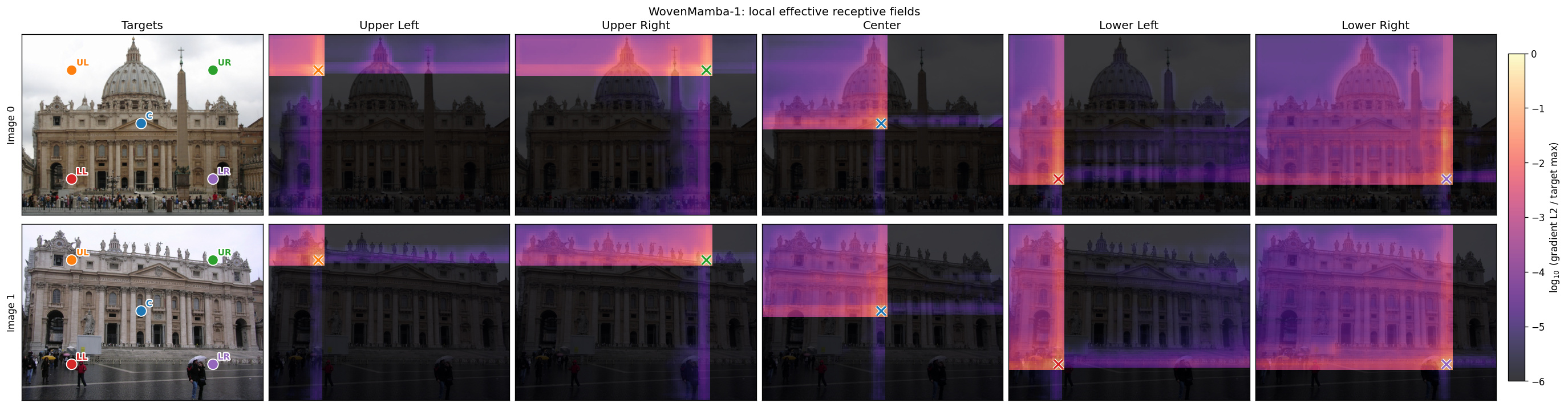}
    \par\vspace{0.4em}
    \includegraphics[width=\linewidth]{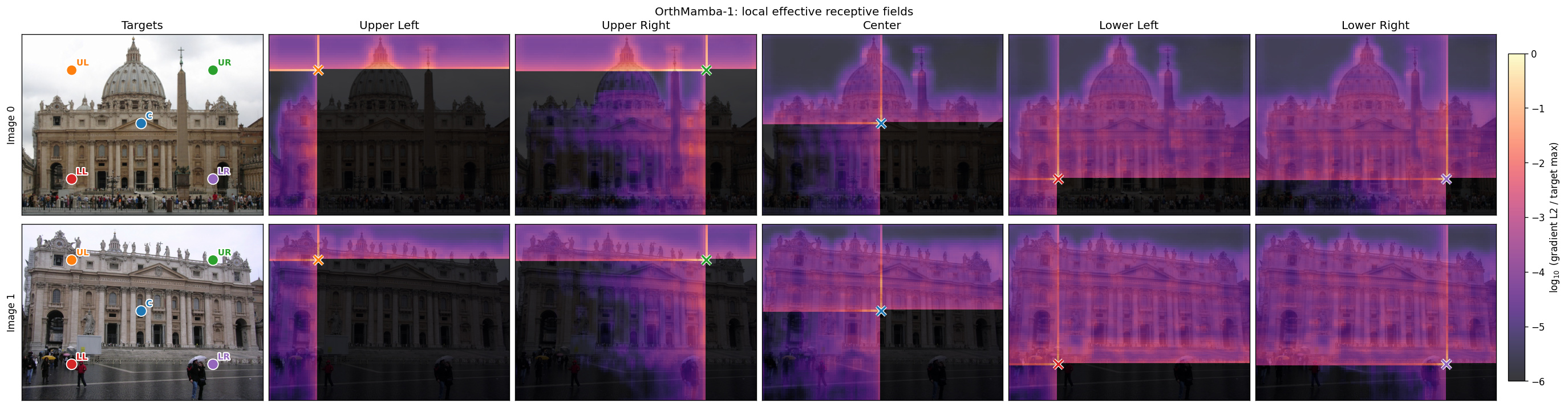}
    \caption{\textbf{Local ERFs of trained intra-image context aggregation blocks.} WovenMamba-1 (top) and OrthMamba-1 (bottom) for image pair 1. Brighter colors indicate greater relative sensitivity to the block input.}
    \label{fig:trained_erf}
\end{figure}

\begin{figure}[htbp]
    \centering
    \includegraphics[width=\linewidth]{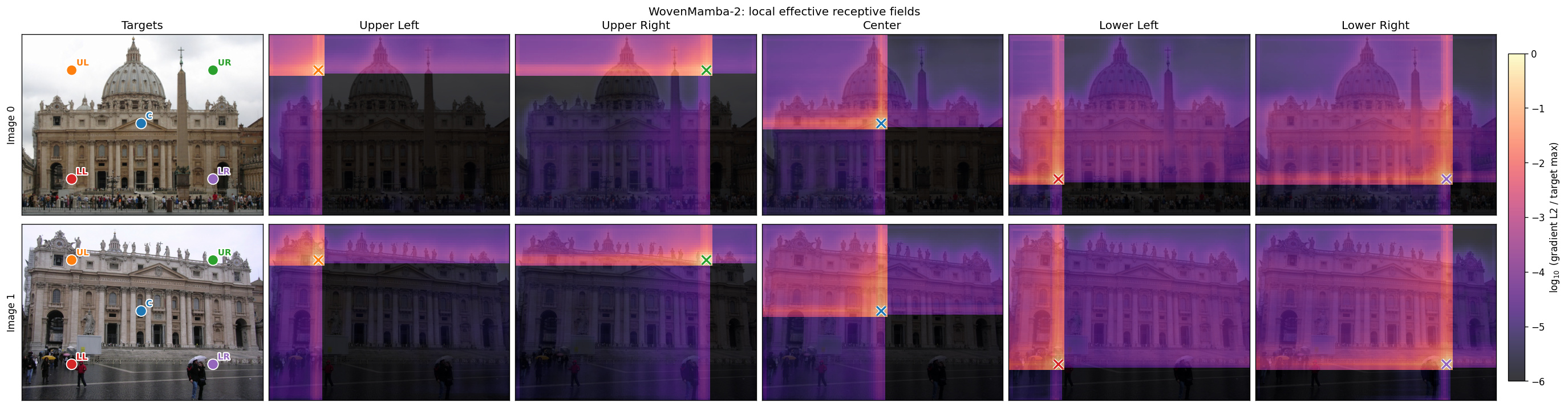}
    \par\vspace{0.4em}
    \includegraphics[width=\linewidth]{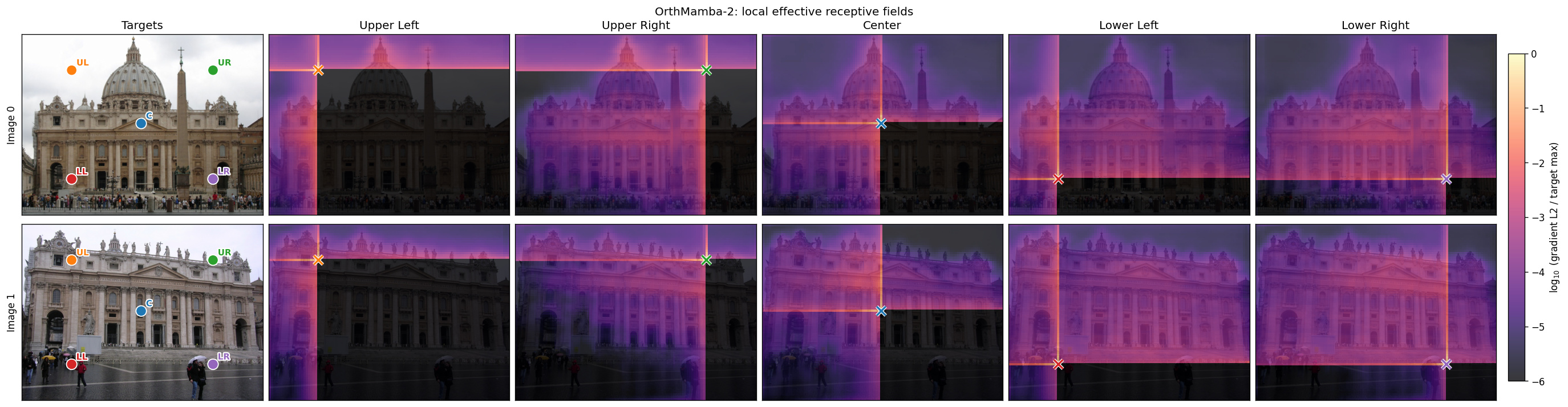}
    \caption{\textbf{Local ERFs of trained intra-image context aggregation blocks.} WovenMamba-2 (top) and OrthMamba-2 (bottom) for image pair 1. Panel layout and color normalization follow the protocol in Appendix~\ref{app:woven_erf}.}
    \label{fig:trained_erf_0015_block2}
\end{figure}

\begin{figure}[p]
    \centering
    \includegraphics[width=\linewidth]{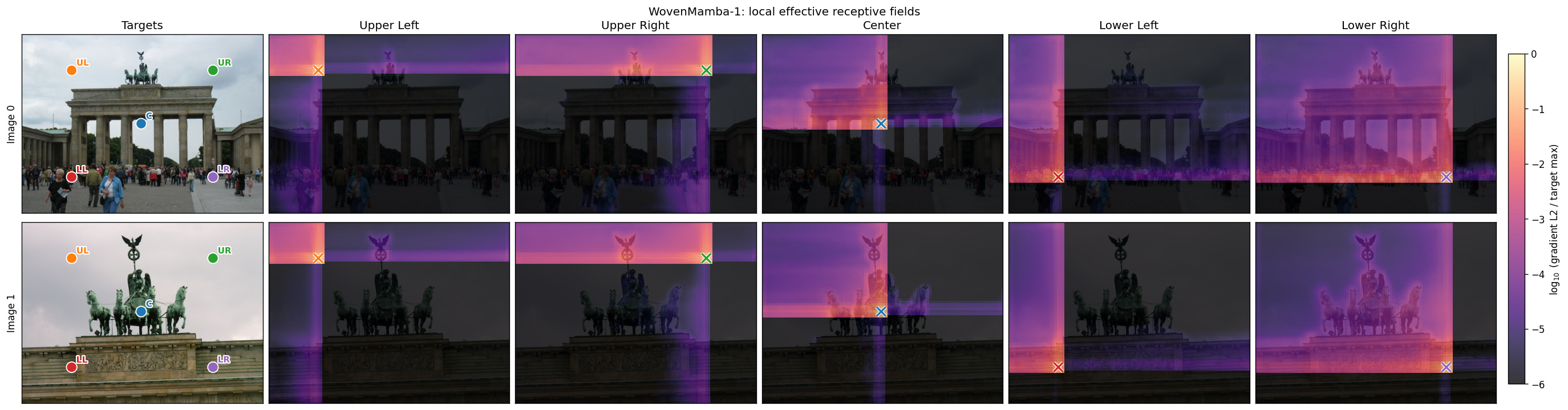}
    \par\vspace{0.4em}
    \includegraphics[width=\linewidth]{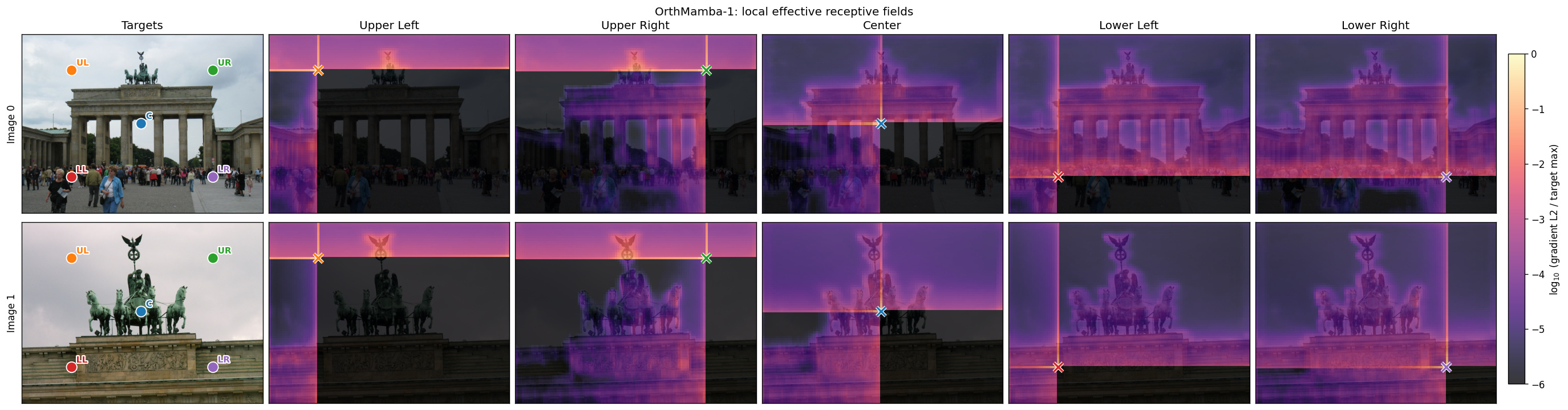}
    \caption{\textbf{Local ERFs of trained intra-image context aggregation blocks.} WovenMamba-1 (top) and OrthMamba-1 (bottom) for image pair 2. Panel layout and color normalization follow the protocol in Appendix~\ref{app:woven_erf}.}
    \label{fig:trained_erf_0022_block1}
\end{figure}

\begin{figure}[p]
    \centering
    \includegraphics[width=\linewidth]{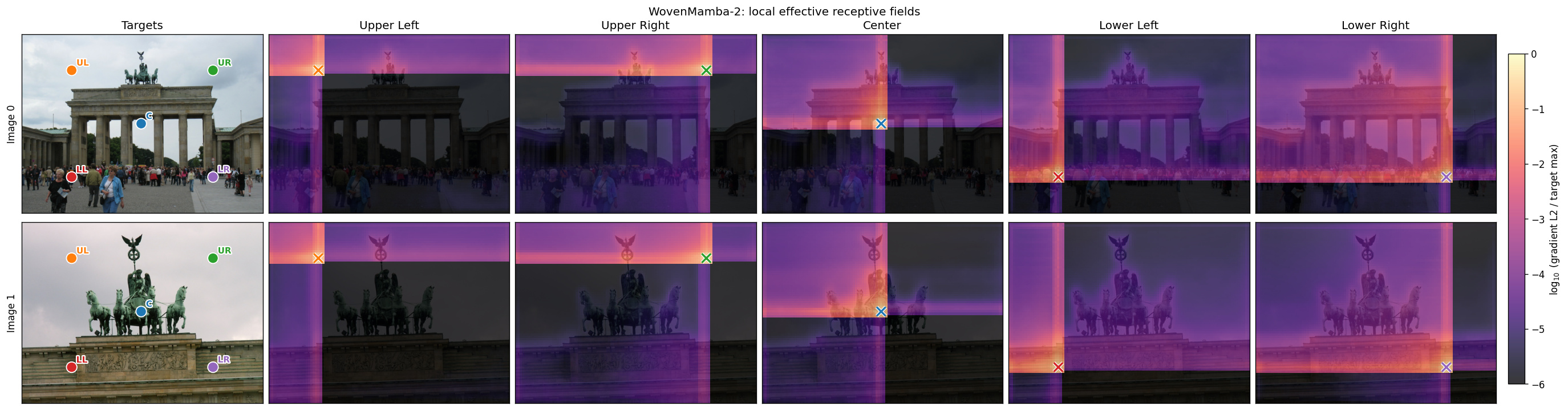}
    \par\vspace{0.4em}
    \includegraphics[width=\linewidth]{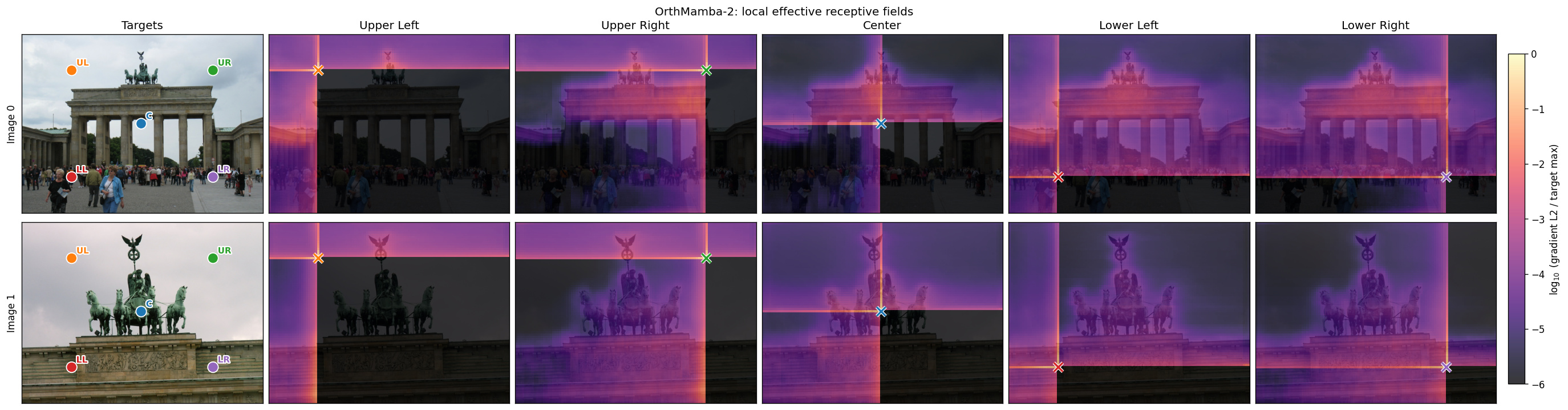}
    \caption{\textbf{Local ERFs of trained intra-image context aggregation blocks.} WovenMamba-2 (top) and OrthMamba-2 (bottom) for image pair 2. Panel layout and color normalization follow the protocol in Appendix~\ref{app:woven_erf}.}
    \label{fig:trained_erf_0022_block2}
\end{figure}

Figure~\ref{fig:trained_erf} compares the first spatial block of WovenMamba and OrthMamba for image pair 1. Figure~\ref{fig:trained_erf_0015_block2} shows their second spatial block for the same pair, and Figures~\ref{fig:trained_erf_0022_block1} and~\ref{fig:trained_erf_0022_block2} show both blocks for image pair 2. The visualizations use the learned parameters of the respective trained models, including their nonlinearities, gating, and normalization. Each operator visualization has two image rows, with target locations in the first column and the corresponding ERFs in the remaining five columns.

\paragraph{Computation.}
Let $\mathbf{X}$ denote the input feature map to spatial block $b$, and let $\Delta_b(\mathbf{X})$ denote its residual update before adding the identity connection. For a target position $t$, we compute the channel-wise gradient magnitude at each source position $p$:
\begin{equation}
    E_t(p)
    =
    \left\|
    \frac{\partial\|\Delta_b(\mathbf{X})_t\|_2}
         {\partial\mathbf{X}_p}
    \right\|_2.
\end{equation}
For valid positions $V$, the displayed value is
\begin{equation}
    \widehat{E}_t(p)
    =
    \log_{10}\!\left(
    \max\!\left(
    \frac{E_t(p)}{\max_{q\in V}E_t(q)},10^{-6}
    \right)\right).
\end{equation}
Thus, all maps use the same color range $[-6,0]$ and show relative spatial influence rather than absolute gradient magnitudes. We select five valid coarse-grid targets near the upper left, upper right, center, lower left, and lower right, and mask padded positions. Gradients are taken with respect to the current block input, detached from preceding layers. In particular, the second-block maps describe local sensitivity at a later contextualized representation, rather than gradients through the entire matcher to the input image.

\paragraph{Observed spatial patterns.}
The contrast between WovenMamba's coherent two-dimensional neighborhood responses and OrthMamba's more pronounced row- and column-aligned responses recurs across both image pairs and spatial blocks shown.

\clearpage
\section{AnchorCrossAttention details}
\label{app:aca_details}

\paragraph{Window definition.}
Let $(u_m,v_m)$ denote the top-left coordinate of the $m$-th window on the coarse feature grid. Its spatial support is
\begin{equation}
    \mathcal{W}_m
    =
    \{(u,v)\in\mathbb{Z}^2:
    u_m\le u<u_m+w,\quad v_m\le v<v_m+w\}.
\end{equation}
Window origins are spaced $w/2$ apart along both axes. When needed, we zero-pad the bottom and right boundaries to align the grid with the stride and mask padded positions during computation.

\paragraph{Anchor construction.}
Omitting the image superscript, let $\mathbf{x}_{m}(u,v)$ denote the feature at $(u,v)$ in window $\mathcal{W}_m$ of $\widetilde{\mathbf{F}}_c$. Within each window, we normalize each feature channel using its spatial mean and variance over valid positions, adding $10^{-5}$ to the variance for numerical stability. Soft assignments are computed from these normalized features, while both raw and normalized features contribute to the anchor descriptor. Let $S_{m,k}=\sum_{(u,v)\in\mathcal{W}_m}P_{m,k}(u,v)$ and $\varepsilon=10^{-6}$. We divide the average weighted features by the average assignment weight, clamped below at $\varepsilon$ for numerical stability. The appearance summaries are
\begin{equation}
\begin{aligned}
    \mathbf{f}^{\mathrm{raw}}_{m,k}
    &=
    \frac{
    \frac{1}{w^2}\sum_{(u,v)\in\mathcal{W}_m}
    P_{m,k}(u,v)\mathbf{x}_{m}(u,v)
    }{\max\!\left(\frac{S_{m,k}}{w^2},\varepsilon\right)},\\
    \mathbf{f}^{\mathrm{norm}}_{m,k}
    &=
    \frac{
    \frac{1}{w^2}\sum_{(u,v)\in\mathcal{W}_m}
    P_{m,k}(u,v)\bar{\mathbf{x}}_{m}(u,v)
    }{\max\!\left(\frac{S_{m,k}}{w^2},\varepsilon\right)}.
\end{aligned}
\end{equation}
To complement the pooled appearance features, we encode the spatial distribution of each anchor's assignments within its window. The assignment-weighted center and covariance summarize spatial location and spread, while the assignment mass measures the total membership weight. Using local coordinates $\mathbf{p}_{m}(u,v)\in[-1,1]^2$ and $Z_{m,k}=S_{m,k}+\varepsilon$, we compute the center and covariance as follows:
\begin{align}
    \boldsymbol{\mu}_{m,k}
    &=
    \frac{
    \sum_{(u,v)\in\mathcal{W}_m}P_{m,k}(u,v)\mathbf{p}_{m}(u,v)
    }{Z_{m,k}},\\
    \Sigma_{m,k}
    &=
    \frac{
    \sum_{(u,v)\in\mathcal{W}_m}P_{m,k}(u,v)
    \mathbf{d}_{m,k}(u,v)\mathbf{d}_{m,k}(u,v)^\top
    }{Z_{m,k}},
\end{align}
where $\mathbf{d}_{m,k}(u,v)=\mathbf{p}_{m}(u,v)-\boldsymbol{\mu}_{m,k}$.
The six geometry statistics are
\begin{equation}
    \mathbf{g}_{m,k}
    =
    [
    \boldsymbol{\mu}_{m,k},
    \Sigma^{xx}_{m,k},
    \Sigma^{yy}_{m,k},
    \Sigma^{xy}_{m,k},
    \log Z_{m,k}
    ].
\end{equation}
A learned linear layer maps these statistics to $\mathbf{f}^{\mathrm{geo}}_{m,k}\in\mathbb{R}^{C_g}$. We then form
\begin{equation}
    \mathbf{a}_{m,k}
    =
    \operatorname{Linear}_{\mathrm{anc}}
    \left(
    [
    \mathbf{f}^{\mathrm{raw}}_{m,k},
    \mathbf{f}^{\mathrm{norm}}_{m,k},
    \mathbf{f}^{\mathrm{geo}}_{m,k}
    ]
    \right).
\end{equation}
Clustering parameters are shared across windows and images within a block; different ACA blocks have separate parameters. Within cross-attention, each anchor descriptor is normalized with RMSNorm before the learned query, key, and value projections.

\paragraph{Overlap weighting.}
A token can have different assignment distributions in different overlapping windows. For overlap fusion after anchor-to-feature projection, we use assignment entropy as a cue for learning their relative contributions. For each valid token position, let $e_m(u,v)$ denote assignment entropy, $c_m(u,v)$ the unnormalized window weight, and $\lambda_m(u,v)$ its normalized counterpart. With the same $\varepsilon=10^{-6}$, we compute
\begin{align}
    e_{m}(u,v)
    &=
    -\sum_{k=0}^{K-1}
    P_{m,k}(u,v)
    \log(P_{m,k}(u,v)+\varepsilon),\\
    c_{m}(u,v)
    &=
    \operatorname{sigmoid}
    \bigl(\alpha_c e_{m}(u,v)+\beta_c\bigr),\\
    \lambda_{m}(u,v)
    &=
    \frac{c_{m}(u,v)}
    {\max\!\left(
    \sum_{m'\in\mathcal{I}(u,v)}c_{m'}(u,v),
    \varepsilon\right)}.
\end{align}
The scalars $\alpha_c$ and $\beta_c$ are learned. This parameterization learns how entropy affects the contribution of each window; it does not impose a fixed preference for lower-entropy assignments.

\paragraph{Component analysis.}
Table~\ref{tab:aca_components} evaluates the ACA components on MegaDepth-1500 while keeping the remaining configuration fixed. Uniform assignments, removing geometry features, and uniform overlap weighting each reduce AUC@5$^\circ$, supporting learned assignments, spatial information, and adaptive overlap weighting under the evaluated setting. Joint attention across all anchor IDs improves AUC@5$^\circ$ by 0.2 points at a cost of 1.0 ms in full-matcher runtime. We therefore use attention within each anchor-ID group as an accuracy--computation trade-off.

\begin{table}[htbp]
\centering
\caption{\textbf{ACA component ablations on MegaDepth-1500.} Runtime is measured for the full matcher.}
\label{tab:aca_components}
\begin{tabular}{lcccc}
\toprule
Variant & AUC@5$^\circ$ & AUC@10$^\circ$ & AUC@20$^\circ$ & Runtime (ms) \\
\midrule
Default ACA & 66.1 & 78.9 & 87.7 & 35.4 \\
Without geometry features & 65.8 & 78.7 & 87.6 & 35.1 \\
Joint attention across anchor IDs & 66.3 & 79.1 & 87.8 & 36.4 \\
Uniform anchor assignments & 65.7 & 78.7 & 87.4 & 34.7 \\
Uniform overlap weights & 65.7 & 78.8 & 87.7 & 34.8 \\
\bottomrule
\end{tabular}
\end{table}

\paragraph{Anchor assignment behavior.}
Table~\ref{tab:anchor_assignment_stats} summarizes the assignment probabilities in the first (ACA-1) and second (ACA-2) ACA blocks on MegaDepth-1500. We pool valid token--window entries from both images of all evaluation pairs; occurrences in overlapping windows are counted separately. ACA-1 has nearly balanced mean assignments, while ACA-2 has less uniform means and larger population variances for all four anchors, indicating greater variation in the assignment probabilities. Shared clustering parameters provide a common assignment basis across images. Together with the assignment maps for two image pairs in Appendix~\ref{app:aca_visualizations}, these statistics suggest recurring spatial preferences that become more distinct in ACA-2. These observations do not establish fixed semantic categories.

\begin{table}[htbp]
\centering
\caption{\textbf{Anchor assignment statistics on MegaDepth-1500.} Mean and population variance of assignment probabilities in the two ACA blocks.}
\label{tab:anchor_assignment_stats}
\begin{tabular}{ccccc}
\toprule
 & \multicolumn{2}{c}{ACA-1} & \multicolumn{2}{c}{ACA-2} \\
\cmidrule(lr){2-3}\cmidrule(lr){4-5}
Anchor $k$ & Mean & Variance & Mean & Variance \\
\midrule
0 & 0.251 & 0.008 & 0.217 & 0.012 \\
1 & 0.249 & 0.007 & 0.267 & 0.042 \\
2 & 0.253 & 0.008 & 0.232 & 0.020 \\
3 & 0.247 & 0.006 & 0.284 & 0.047 \\
\bottomrule
\end{tabular}
\end{table}

\paragraph{Sensitivity to anchor count and window size.}
Table~\ref{tab:aca_sensitivity} reports sweeps over $K\in\{2,3,4,5,6\}$ with $w=8$ and $w\in\{4,6,8,10,12\}$ with $K=4$, retaining 50\% overlap. Among the tested anchor counts, $K=4$ achieves the highest AUC@5$^\circ$; increasing $K$ beyond 4 adds runtime without improving accuracy. A smaller window of size 4 achieves 66.3 AUC@5$^\circ$ at 41.3 ms, compared with 66.1 at 35.4 ms for the default window size 8. The default configuration balances accuracy and computation within the tested range.

\begin{table}[htbp]
\centering
\caption{\textbf{Sensitivity to anchor count and window size on MegaDepth-1500.} Runtime is measured for the full matcher.}
\label{tab:aca_sensitivity}
\begin{tabular}{lcccc}
\toprule
Setting & AUC@5$^\circ$ & AUC@10$^\circ$ & AUC@20$^\circ$ & Runtime (ms) \\
\midrule
\multicolumn{5}{l}{\textit{(a) Anchor count, with $w=8$}} \\
$K=2$ & 65.5 & 78.4 & 87.3 & 32.9 \\
$K=3$ & 65.7 & 78.6 & 87.4 & 33.8 \\
$K=4$ (default) & 66.1 & 78.9 & 87.7 & 35.4 \\
$K=5$ & 65.7 & 78.8 & 87.6 & 36.1 \\
$K=6$ & 65.7 & 78.6 & 87.5 & 37.1 \\
\midrule
\multicolumn{5}{l}{\textit{(b) Window size, with $K=4$}} \\
$w=4$ & 66.3 & 79.1 & 87.7 & 41.3 \\
$w=6$ & 66.0 & 78.8 & 87.6 & 38.3 \\
$w=8$ (default) & 66.1 & 78.9 & 87.7 & 35.4 \\
$w=10$ & 66.0 & 78.7 & 87.5 & 35.6 \\
$w=12$ & 65.8 & 78.5 & 87.3 & 35.7 \\
\bottomrule
\end{tabular}
\end{table}

\subsection{Anchor assignment visualizations and analysis}
\label{app:aca_visualizations}

\begin{figure}[htbp]
    \centering
    \includegraphics[width=0.98\linewidth]{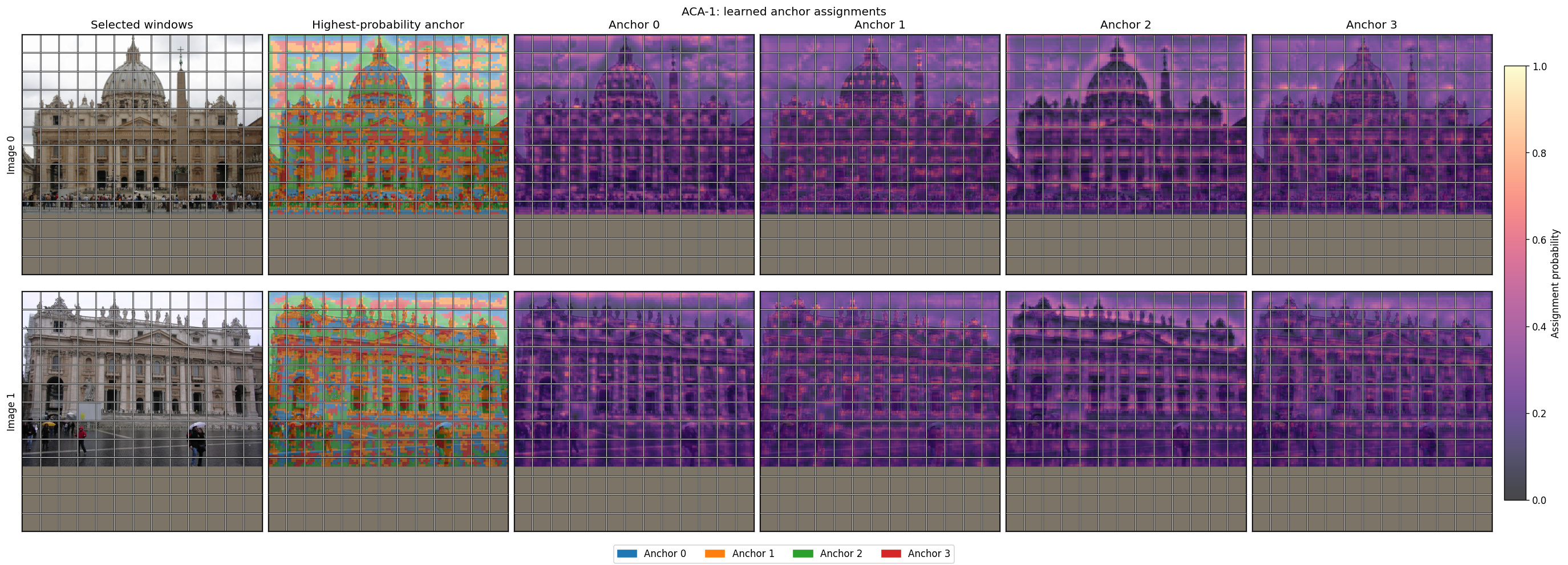}
    \par\vspace{0.4em}
    \includegraphics[width=0.98\linewidth]{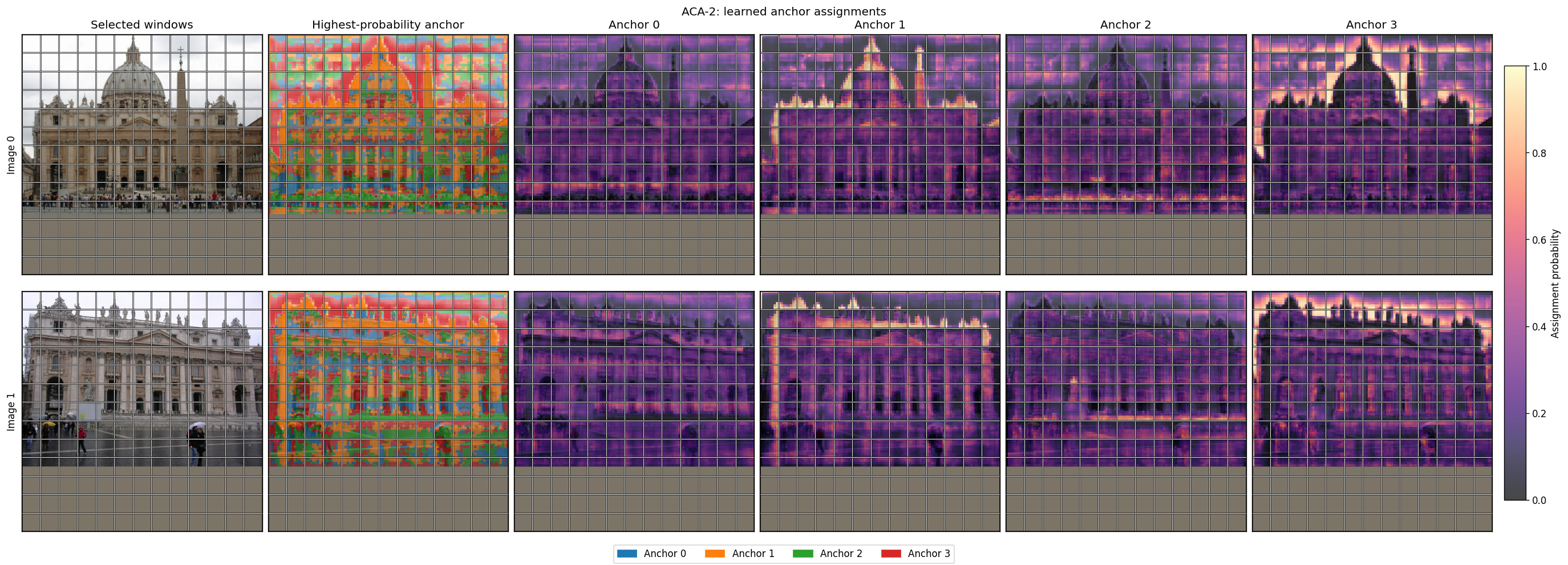}
    \caption{\textbf{Learned anchor assignments in AnchorCrossAttention.} ACA-1 (top) and ACA-2 (bottom) for image pair 1, window offset $(0,0)$. Brighter heatmap colors indicate higher assignment probabilities.}
    \label{fig:aca_0015_lt}
\end{figure}

\begin{figure}[p]
    \centering
    \includegraphics[width=0.98\linewidth]{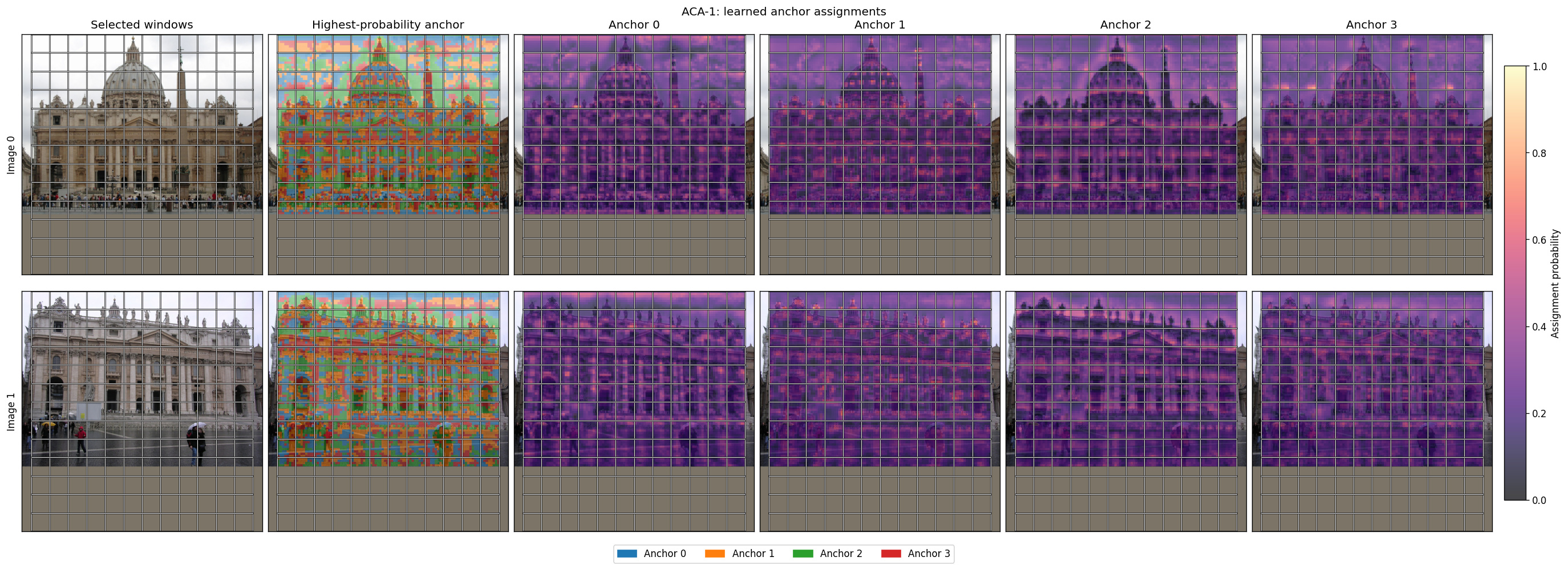}
    \par\vspace{0.4em}
    \includegraphics[width=0.98\linewidth]{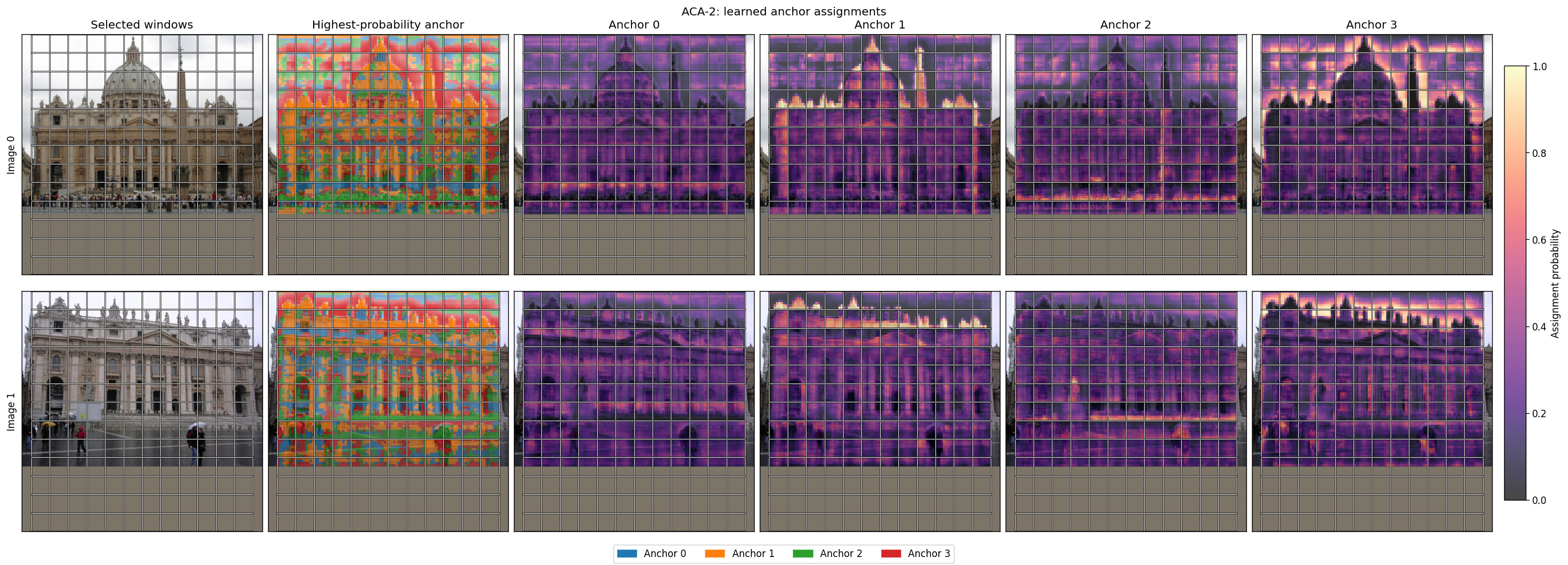}
    \caption{\textbf{Learned anchor assignments in AnchorCrossAttention.} ACA-1 (top) and ACA-2 (bottom) for image pair 1, window offset $(0,4)$. See Appendix~\ref{app:aca_visualizations} for visualization details.}
    \label{fig:aca_0015_rt}
\end{figure}

\begin{figure}[p]
    \centering
    \includegraphics[width=0.98\linewidth]{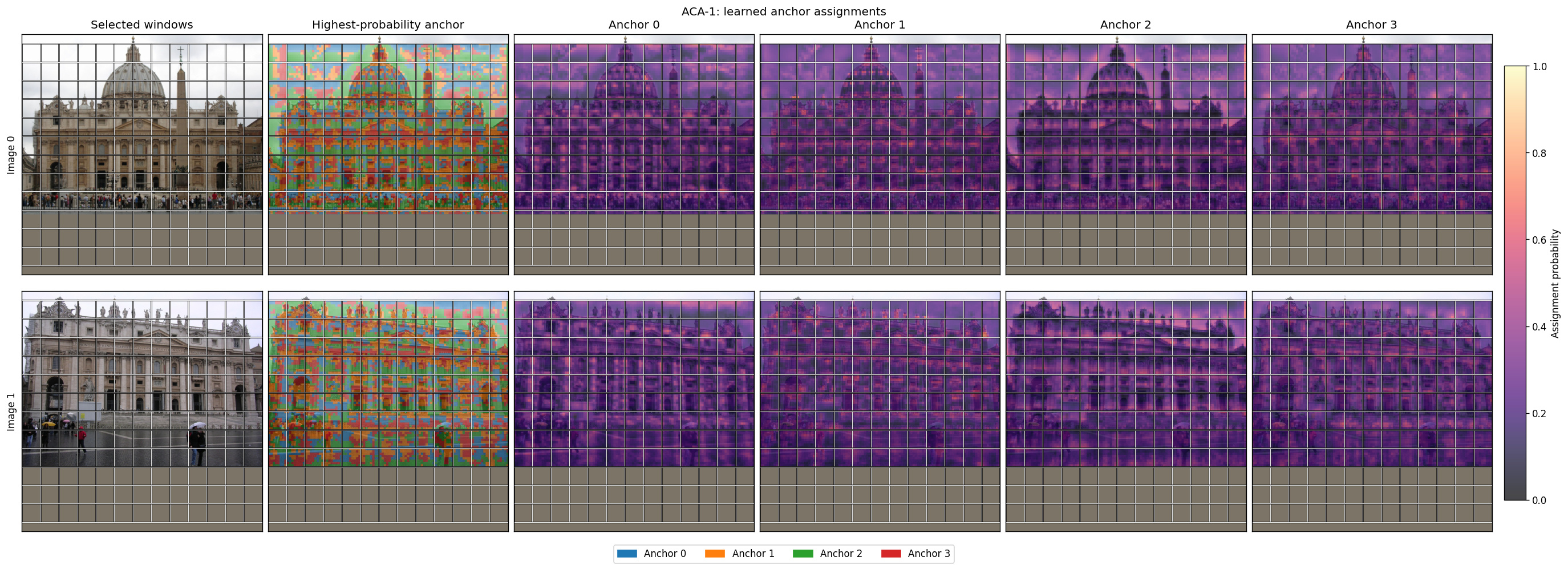}
    \par\vspace{0.4em}
    \includegraphics[width=0.98\linewidth]{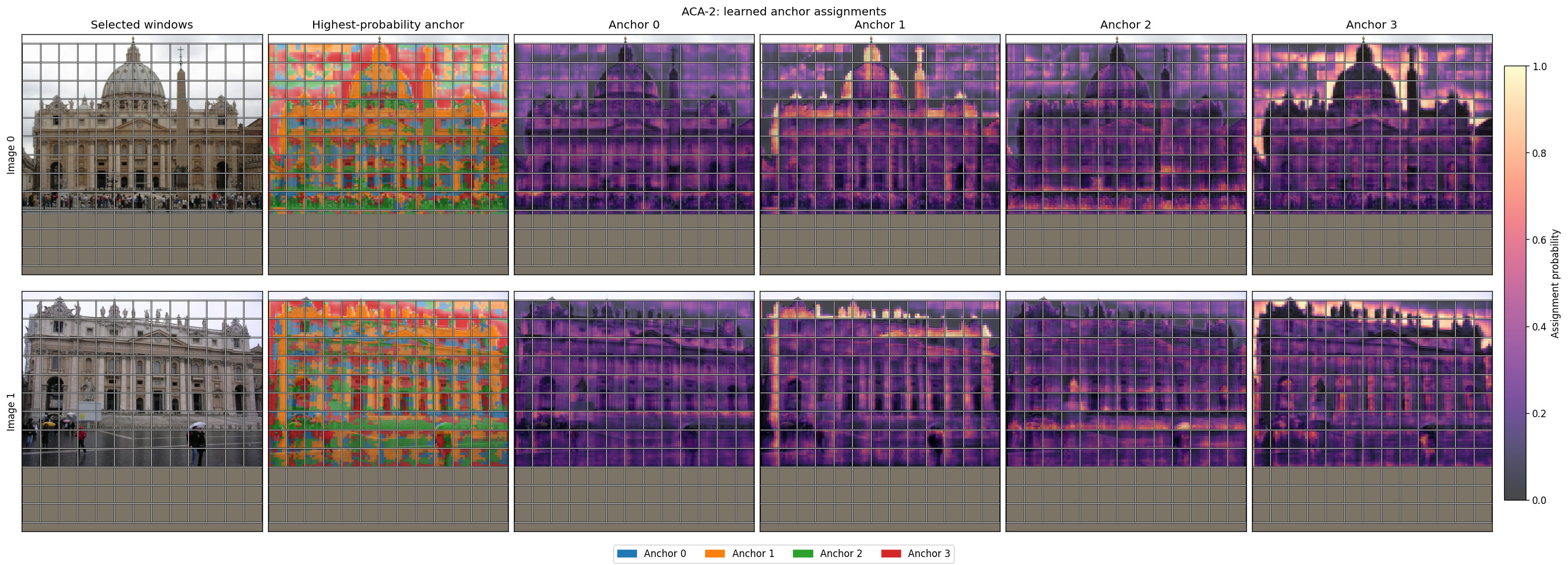}
    \caption{\textbf{Learned anchor assignments in AnchorCrossAttention.} ACA-1 (top) and ACA-2 (bottom) for image pair 1, window offset $(4,0)$. See Appendix~\ref{app:aca_visualizations} for visualization details.}
    \label{fig:aca_0015_lb}
\end{figure}

\begin{figure}[p]
    \centering
    \includegraphics[width=0.98\linewidth]{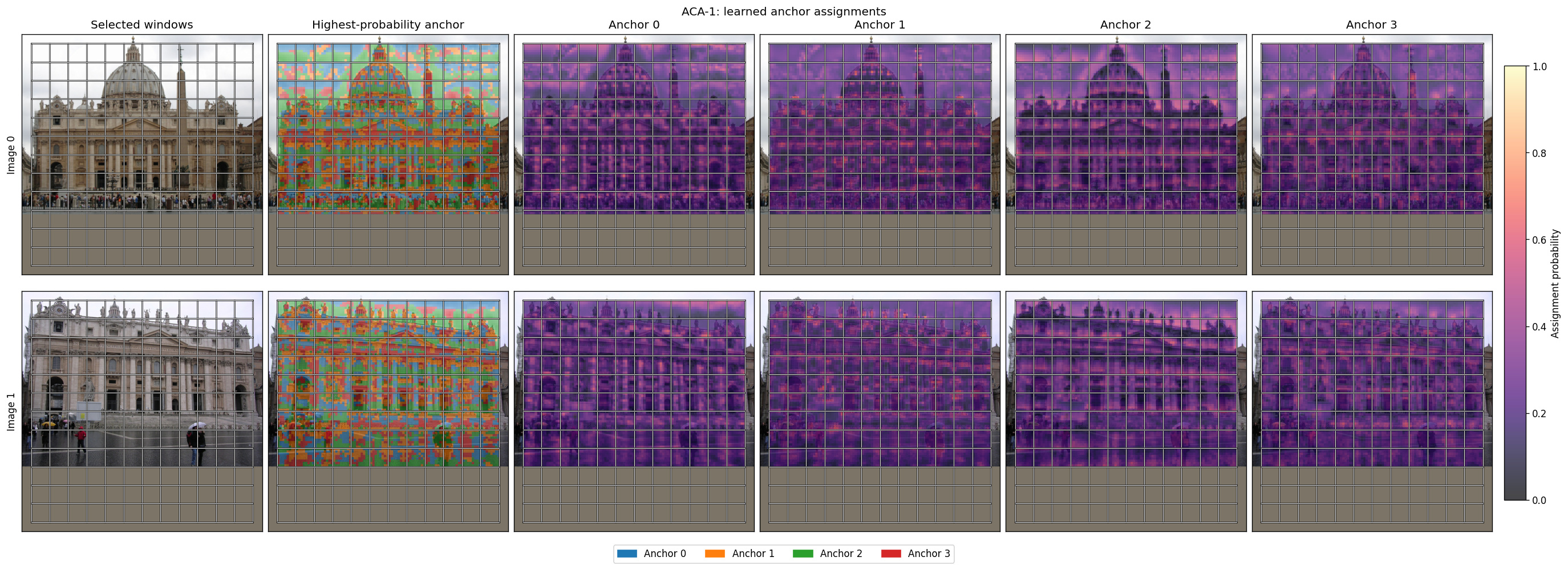}
    \par\vspace{0.4em}
    \includegraphics[width=0.98\linewidth]{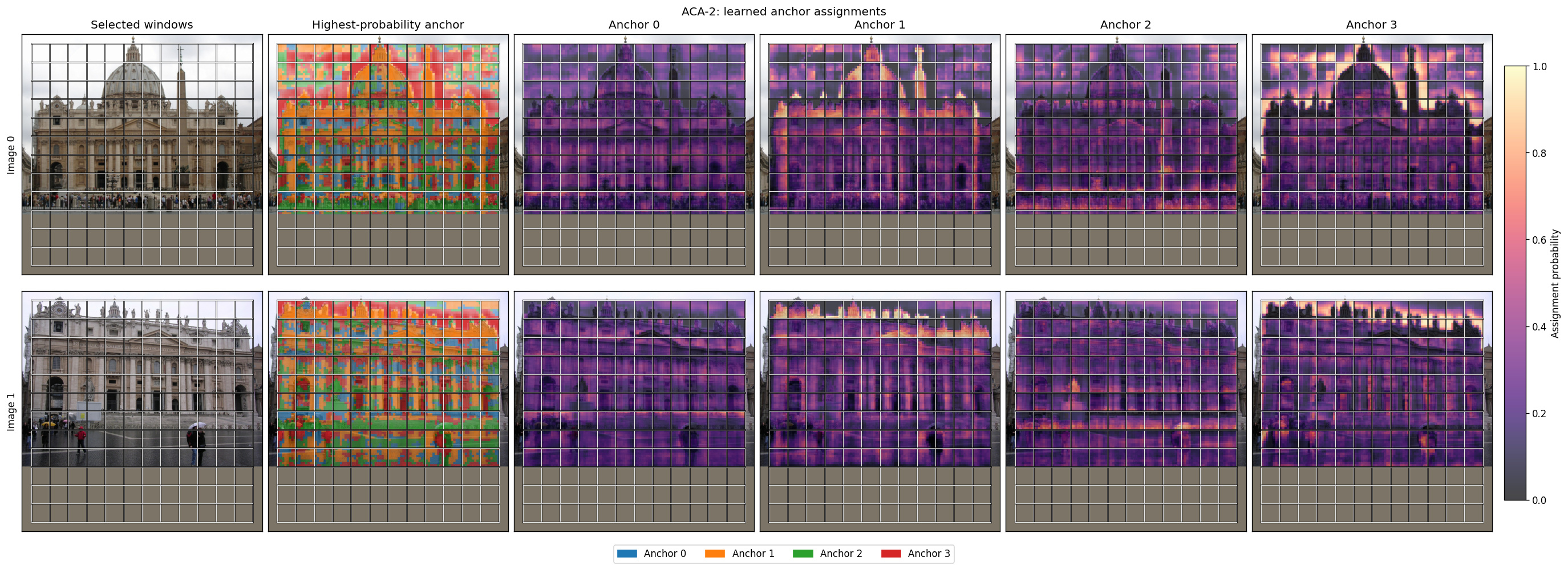}
    \caption{\textbf{Learned anchor assignments in AnchorCrossAttention.} ACA-1 (top) and ACA-2 (bottom) for image pair 1, window offset $(4,4)$. See Appendix~\ref{app:aca_visualizations} for visualization details.}
    \label{fig:aca_0015_rb}
\end{figure}

\begin{figure}[p]
    \centering
    \includegraphics[width=0.98\linewidth]{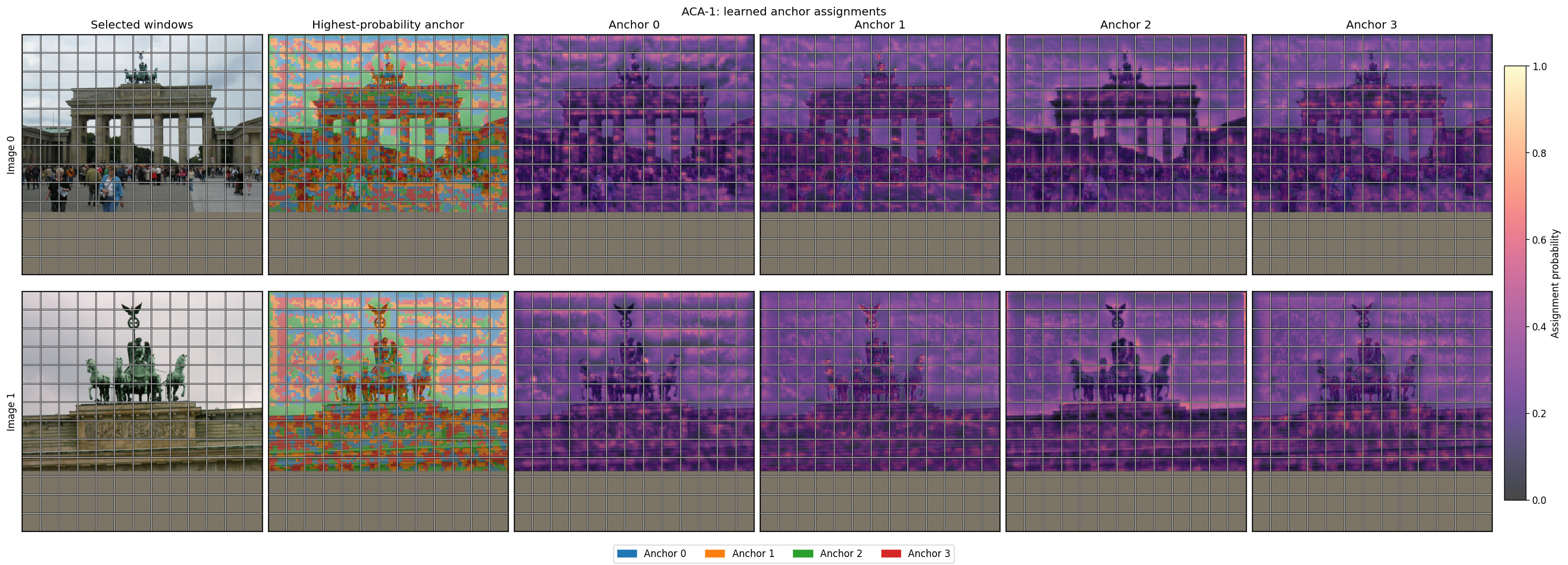}
    \par\vspace{0.4em}
    \includegraphics[width=0.98\linewidth]{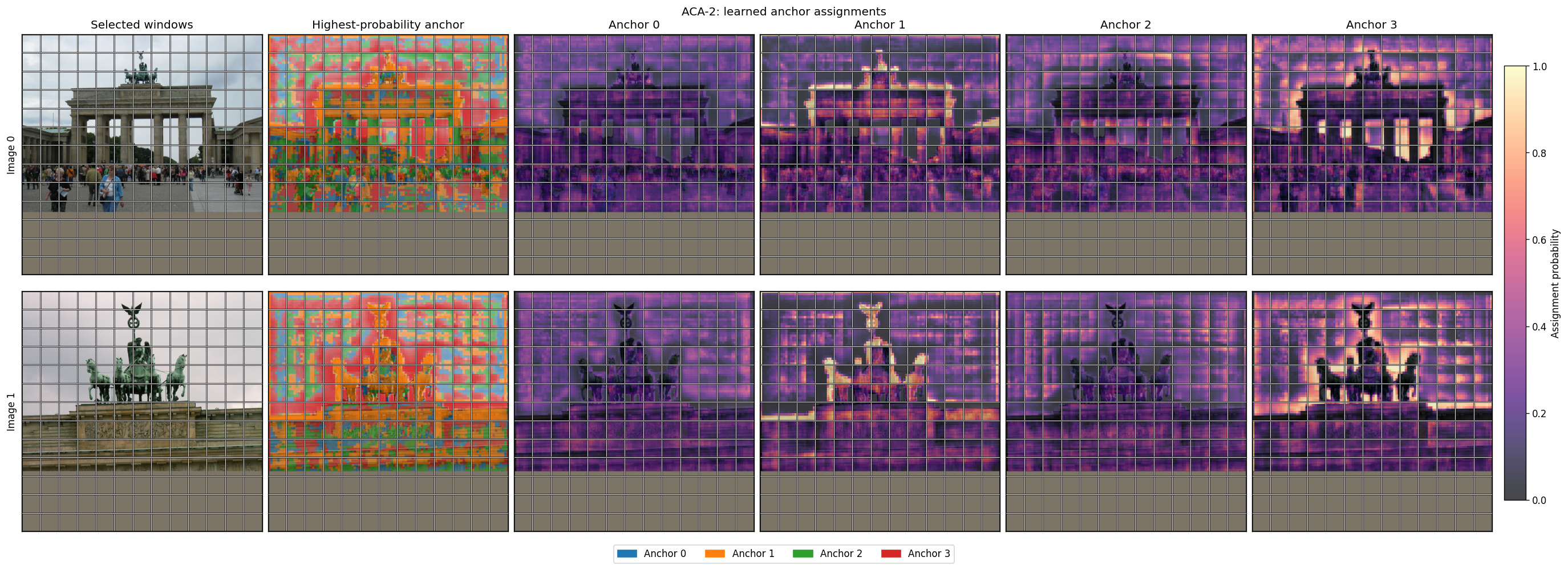}
    \caption{\textbf{Learned anchor assignments in AnchorCrossAttention.} ACA-1 (top) and ACA-2 (bottom) for image pair 2, window offset $(0,0)$. See Appendix~\ref{app:aca_visualizations} for visualization details.}
    \label{fig:aca_0022_lt}
\end{figure}

\begin{figure}[p]
    \centering
    \includegraphics[width=0.98\linewidth]{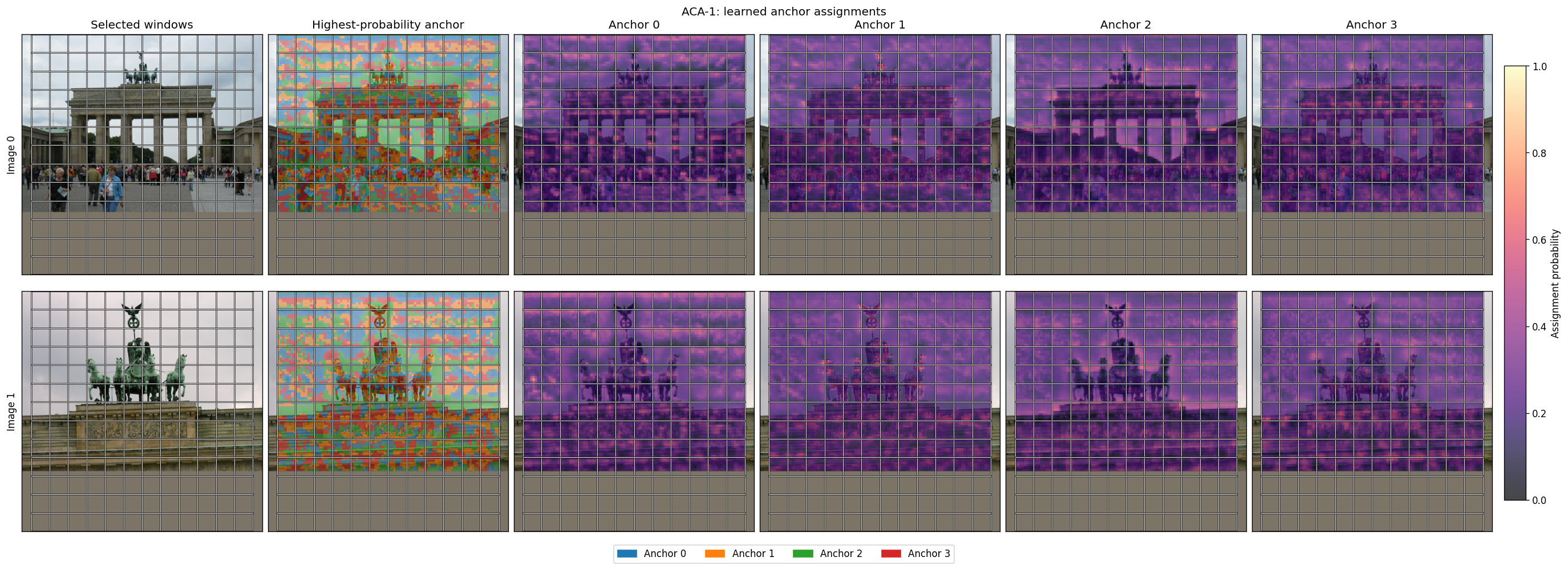}
    \par\vspace{0.4em}
    \includegraphics[width=0.98\linewidth]{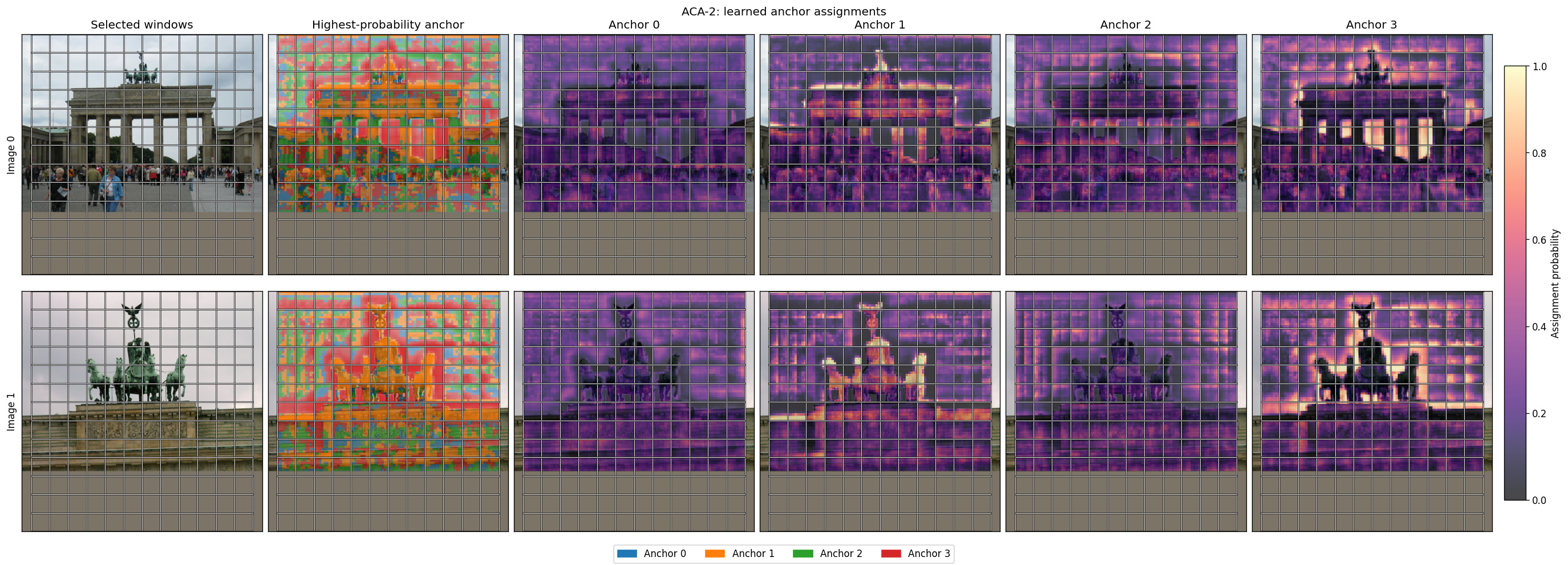}
    \caption{\textbf{Learned anchor assignments in AnchorCrossAttention.} ACA-1 (top) and ACA-2 (bottom) for image pair 2, window offset $(0,4)$. See Appendix~\ref{app:aca_visualizations} for visualization details.}
    \label{fig:aca_0022_rt}
\end{figure}

\begin{figure}[p]
    \centering
    \includegraphics[width=0.98\linewidth]{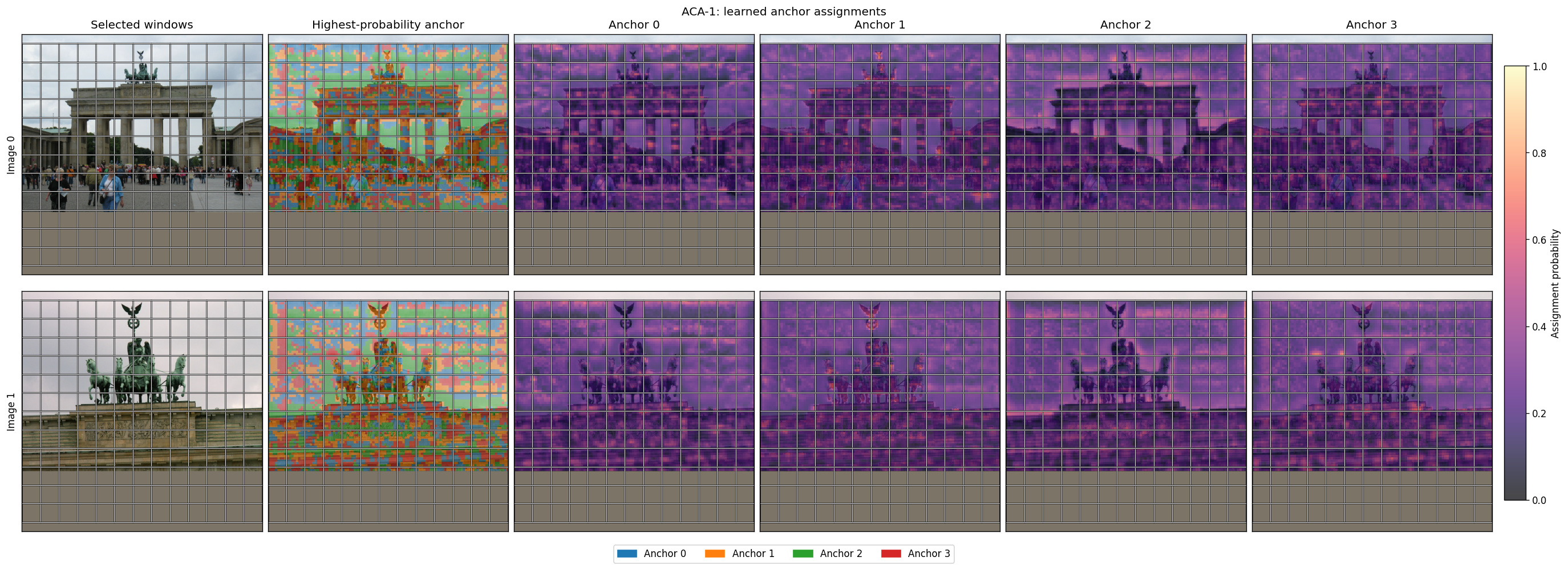}
    \par\vspace{0.4em}
    \includegraphics[width=0.98\linewidth]{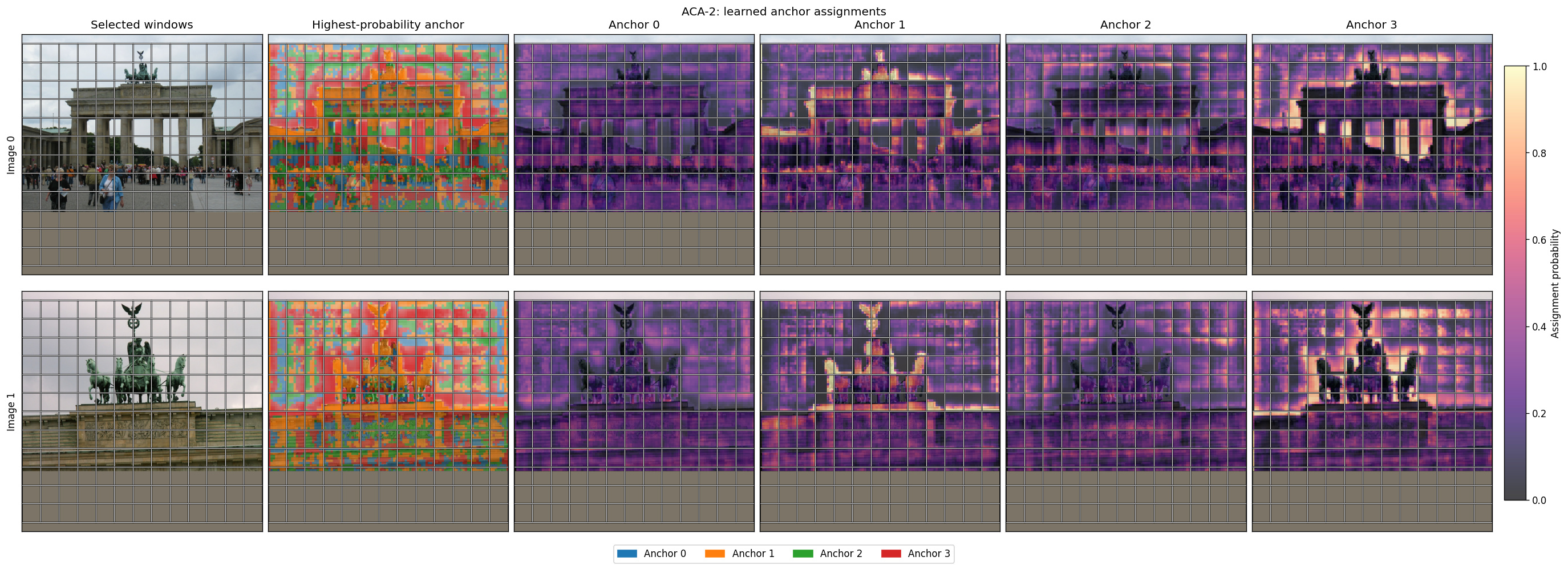}
    \caption{\textbf{Learned anchor assignments in AnchorCrossAttention.} ACA-1 (top) and ACA-2 (bottom) for image pair 2, window offset $(4,0)$. See Appendix~\ref{app:aca_visualizations} for visualization details.}
    \label{fig:aca_0022_lb}
\end{figure}

\begin{figure}[p]
    \centering
    \includegraphics[width=0.98\linewidth]{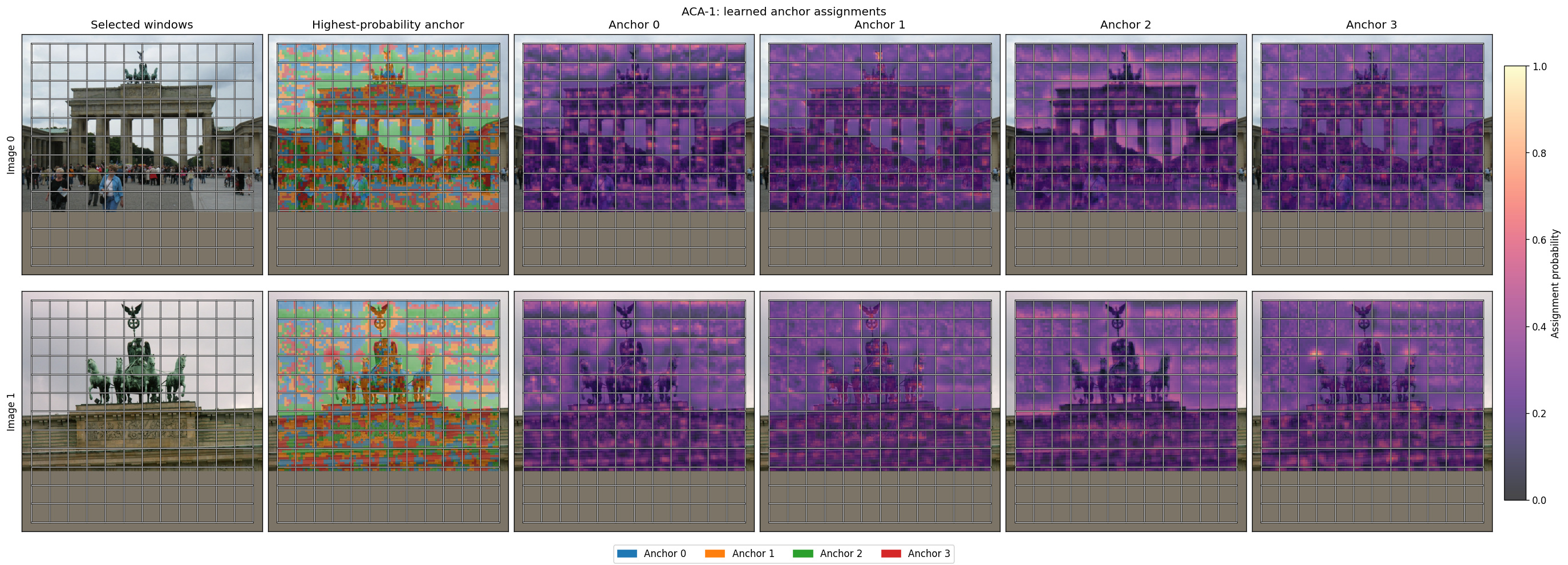}
    \par\vspace{0.4em}
    \includegraphics[width=0.98\linewidth]{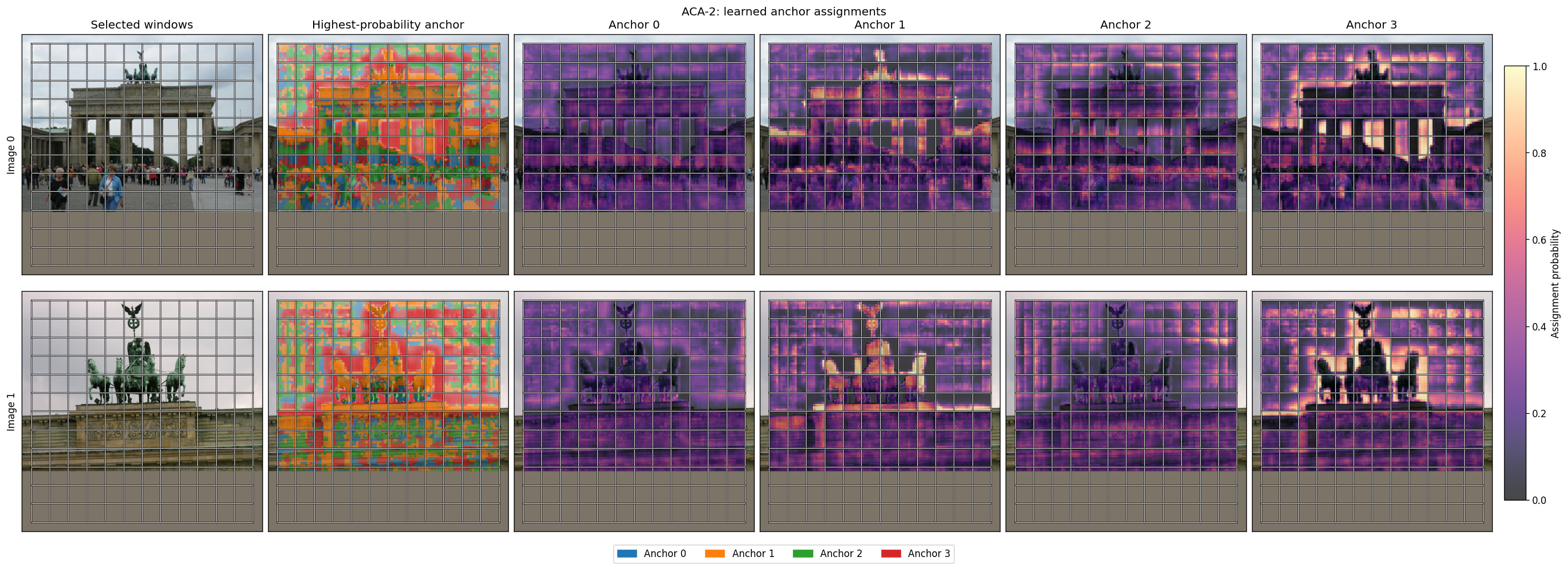}
    \caption{\textbf{Learned anchor assignments in AnchorCrossAttention.} ACA-1 (top) and ACA-2 (bottom) for image pair 2, window offset $(4,4)$. See Appendix~\ref{app:aca_visualizations} for visualization details.}
    \label{fig:aca_0022_rb}
\end{figure}

Figures~\ref{fig:aca_0015_lt}--\ref{fig:aca_0015_rb} show image pair 1; Figures~\ref{fig:aca_0022_lt}--\ref{fig:aca_0022_rb} show image pair 2 from a different scene. Each ACA block is shown as two image rows, with columns for selected windows, the highest-probability anchor, and assignment probabilities for anchors $0$--$3$.

With $8\times8$ windows and stride 4, the windows form four interleaved subsets with origin offsets $(0,0)$, $(0,4)$, $(4,0)$, and $(4,4)$, measured as (row, column) coordinates on the coarse grid. Each displayed subset has stride 8 on the coarse grid, so its windows do not overlap. We visualize each subset separately to retain its window-specific assignments. The categorical maps show $\operatorname*{arg\,max}_k P_{m,k}(u,v)$, and the heatmaps show the individual assignment probabilities on a common $[0,1]$ scale. Assignment overlays are masked out at padded positions and outside the selected windows, leaving the underlying image visible.

\paragraph{Spatial preferences in the two ACA blocks.}
The maps show recurring anchor preferences across the two views of each pair, with more distinct spatial responses in ACA-2. They complement Table~\ref{tab:anchor_assignment_stats} by showing the spatial distribution of the assignments.

\paragraph{Interpretation for anchor grouping.}
Shared clustering vectors provide a common assignment basis within each ACA block. Different blocks have separate parameters, so an anchor ID need not retain the same role across depth. Table~\ref{tab:aca_components} evaluates the accuracy--computation trade-off of attention within anchor-ID groups.

\clearpage

\section{Repeated-run variability}
\label{app:repeated_runs}

Table~\ref{tab:wam_repeated_runs} reports seven independent WAM training runs on MegaDepth-1500 under the same configuration. 
The run identifiers are arbitrary. The default checkpoint used in the main tables is Run D, whose AUC@5$^\circ$ equals the median AUC@5$^\circ$ over the seven runs; ``std'' denotes sample standard deviation.

\begin{table}[h]
\centering
\caption{\textbf{Repeated-run variability of WAM on MegaDepth-1500.}
Runtime is measured on a single NVIDIA H100 GPU under the protocol in Appendix~\ref{app:implementation}.}
\label{tab:wam_repeated_runs}
\begin{tabular}{lcccc}
\toprule
Run & AUC@5$^\circ$ & AUC@10$^\circ$ & AUC@20$^\circ$ & Runtime (ms) \\
\midrule
A & 66.4 & 78.9 & 87.5 & 34.9 \\
B & 66.3 & 79.0 & 87.7 & 35.0 \\
C & 66.2 & 78.9 & 87.6 & 35.0 \\
D (default) & 66.1 & 78.9 & 87.7 & 35.4 \\
E & 66.1 & 78.8 & 87.7 & 34.8 \\
F & 65.9 & 78.5 & 87.4 & 35.0 \\
G & 65.8 & 78.5 & 87.3 & 35.8 \\
\midrule
Mean $\pm$ std & $66.1{\pm}0.2$ & $78.8{\pm}0.2$ & $87.6{\pm}0.2$ & $35.1{\pm}0.3$ \\
Median & 66.1 & 78.9 & 87.6 & 35.0 \\
Min--max & 65.8--66.4 & 78.5--79.0 & 87.3--87.7 & 34.8--35.8 \\
\bottomrule
\end{tabular}
\end{table}

For context, all non-default component-removal and replacement variants in Tables~\ref{tab:woven_anchor_module}--\ref{tab:abl_inter_image} are below the minimum WAM AUC@5$^\circ$ observed across these seven runs.

\section{Input-resolution scalability}
\label{app:resolution_scaling}

We vary the MegaDepth resize-and-pad resolution and evaluate WAM and JamMa using their respective MegaDepth-trained checkpoints and the same matching thresholds. We report both full-matcher and context-module times across resolutions.

\begin{table}[h]
\centering
\caption{\textbf{Input-resolution scalability on MegaDepth.}
We report pose AUC and runtime when varying the input resize-and-pad resolution. Context time measures WovenAnchorModule for WAM and JointMamba Module for JamMa.}
\label{tab:resolution_scaling}
\setlength{\tabcolsep}{3.2pt}
\begin{tabular}{cclccccc}
\toprule
Input & Coarse & Method &
\multicolumn{3}{c}{Pose AUC $\uparrow$} &
Matcher time & Context time \\
\cmidrule(lr){4-6}
size & size & &
@5$^\circ$ & @10$^\circ$ & @20$^\circ$ &
(ms) & (ms) \\
\midrule
\multirow{2}{*}{480}  & \multirow{2}{*}{$60\times60$}
& WAM   & 55.9 & 71.2 & 82.4 & 15.4 & 6.7 \\
& & JamMa & 53.3 & 68.2 & 79.2 & 16.0 & 2.8 \\
\midrule
\multirow{2}{*}{640}  & \multirow{2}{*}{$80\times80$}
& WAM   & 63.7 & 76.7 & 86.3 & 22.0 & 8.8 \\
& & JamMa & 60.3 & 73.9 & 83.8 & 25.0 & 2.8 \\
\midrule
\multirow{2}{*}{832}  & \multirow{2}{*}{$104\times104$}
& WAM   & 66.1 & 78.9 & 87.7 & 35.4 & 14.0 \\
& & JamMa & 64.1 & 77.4 & 86.5 & 43.8 & 2.9 \\
\midrule
\multirow{2}{*}{1024} & \multirow{2}{*}{$128\times128$}
& WAM   & 67.2 & 79.6 & 88.0 & 52.9 & 20.7 \\
& & JamMa & 65.8 & 78.4 & 87.1 & 64.7 & 3.5 \\
\midrule
\multirow{2}{*}{1152} & \multirow{2}{*}{$144\times144$}
& WAM   & 67.6 & 79.8 & 88.3 & 68.2 & 26.2 \\
& & JamMa & 66.2 & 78.5 & 87.3 & 86.4 & 4.9 \\
\midrule
\multirow{2}{*}{1312} & \multirow{2}{*}{$164\times164$}
& WAM   & 68.0 & 80.1 & 88.4 & 91.7 & 34.1 \\
& & JamMa & 66.2 & 78.6 & 87.3 & 121.0 & 5.8 \\
\bottomrule
\end{tabular}
\end{table}

In Table~\ref{tab:resolution_scaling}, WAM obtains higher pose AUC than JamMa at all tested resolutions, supporting the role-specialized design beyond the default $832\times832$ setting. Full-matcher times reflect the complete implementation, including coarse-matching optimizations; context times isolate the cost of the aggregation modules. WovenAnchorModule has a higher measured context-module runtime than JointMamba at all tested resolutions. AnchorCrossAttention retains quadratic interaction complexity in the number of local windows.

\section{Additional visual localization evaluation}
\label{app:aachen}

\paragraph{Aachen Day-Night v1.1.}
We additionally evaluate WAM on Aachen Day-Night v1.1 \citep{sattler2018benchmarking,zhang2021referencepose} for visual localization using the public HLoc benchmark pipeline \citep{sarlin2019coarse}. Since detailed evaluation settings for some published matcher baselines are not always fully specified, we report results from our own evaluation under a shared pipeline. All methods are evaluated with the same localization settings and the same HLoc-based evaluation code. The results should therefore be interpreted as a controlled comparison under our evaluation setup rather than a reproduction of each method's official leaderboard submission. Following the Aachen Day-Night v1.1 protocol, we report localization recall at $(0.25\,\mathrm{m},2^\circ)$, $(0.5\,\mathrm{m},5^\circ)$, and $(5.0\,\mathrm{m},10^\circ)$.

\begin{table}[h]
\centering
\caption{Visual localization results on Aachen Day-Night v1.1. We report localization recall at $(0.25\,\mathrm{m},2^\circ)$ / $(0.5\,\mathrm{m},5^\circ)$ / $(5.0\,\mathrm{m},10^\circ)$. All methods are evaluated under the same HLoc-based pipeline.}
\label{tab:aachen_v11}
\begin{tabular}{lcc}
\toprule
Method & Day & Night \\
\midrule
LoFTR \citep{sun21loftr} & 88.7 / 95.8 / 98.8 & 76.4 / 91.1 / 97.9 \\
JamMa \citep{lu2025jamma} & 86.4 / 94.5 / 97.6 & 72.8 / 90.1 / 98.4 \\
WAM (Ours) & 87.5 / 94.5 / 98.2 & 74.3 / 91.1 / 97.9 \\
\bottomrule
\end{tabular}
\end{table}

\paragraph{Discussion.}
WAM generally improves over JamMa under this controlled HLoc-based evaluation, but it does not outperform LoFTR on Aachen Day-Night v1.1. One possible reason is the match selection strategy. Like JamMa, WAM uses nearest-neighbor matching, which can produce many-to-one assignments. Aggregating or suppressing such matches may introduce additional sensitivity in the localization pipeline. In contrast, LoFTR uses mutual nearest-neighbor filtering in our evaluated pipeline, which more directly suppresses many-to-one correspondences. Compared with JamMa, which uses a similar nearest-neighbor selection strategy, WAM generally performs better, suggesting that the proposed WovenAnchorModule yields more effective matching features than JointMamba in this setting.

\section{Limitations and broader impact}
\label{app:broader_impact}

\paragraph{Responsible use.}
Improved local feature matching can benefit visual localization, SfM, SLAM, mapping, image alignment, augmented reality, and robotic perception \citep{sattler2018benchmarking,sarlin2019coarse,schoenberger2016sfm,murartal2015orbslam,hartley2004multiple}. Potential negative uses include privacy-invasive localization, surveillance, and reconstruction from imagery without adequate consent. Incorrect matches may also propagate errors to downstream robotic or mapping systems. Responsible deployment should consider privacy constraints, consent for imagery, benchmark limitations, and downstream failure handling.

\paragraph{Limitations.}
The nearest-neighbor-based matching strategy can produce many-to-one correspondences, which may contribute to the localization gap observed on Aachen in Appendix~\ref{app:aachen}. The comparison does not isolate the effect of match filtering; localization performance may depend on both the learned features and how correspondences are filtered and aggregated. Another limitation is computational scaling. WovenAnchorModule has a higher measured context-module runtime than JointMamba at the default resolution (Table~\ref{tab:runtime_breakdown}). Although local anchors reduce the number of interactions relative to dense token-wise cross-attention, AnchorCrossAttention retains quadratic interaction complexity in the number of local windows, limiting scalability at high input resolutions. Future work could combine WAM with stronger mutual filtering, one-to-one assignment, localization-specific match pruning, or more scalable anchor retrieval.

\paragraph{Datasets.}
The work uses existing public computer-vision benchmarks and does not involve new human-subject data collection, crowdsourcing, or participant intervention. The method is not a generative model and does not introduce a new scraped dataset.

\paragraph{Existing assets and licenses.}
Table~\ref{tab:assets_licenses} summarizes the main existing datasets, benchmark codebases, and model components used in this work. 
We use these assets for research evaluation. The source package includes adapted third-party code and evaluation metadata inherited from prior work, with applicable licenses and attribution preserved. Dataset images, depth maps, and third-party checkpoints are not bundled.

\begin{table}[h]
\centering
\caption{\textbf{Existing assets and licenses/terms.}}
\label{tab:assets_licenses}
\setlength{\tabcolsep}{2.2pt}
\begin{tabular}{p{0.15\linewidth}p{0.2\linewidth}p{0.5\linewidth}p{0.08\linewidth}}
\toprule
Asset & Use & License / terms & Citation \\
\midrule
MegaDepth & Training and MegaDepth-1500 relative pose evaluation &
CC BY 4.0 for released depth maps, SfM models, and related derived assets; original Internet images retain their original licenses &
\citep{li18megadepth} \\
ScanNet & Zero-shot ScanNet-1500 relative pose evaluation &
ScanNet Terms of Use for data; MIT license for code &
\citep{dai2017scannet} \\
HPatches & Zero-shot homography evaluation &
MIT license for the HPatches dataset repository; BSD-2-Clause license for the HPatches benchmark/evaluation code; no dataset assets are redistributed &
\citep{balntas17hpatches} \\
Aachen Day-Night v1.1 & Visual localization evaluation &
Official dataset/benchmark terms as specified in the accompanying README files; no dataset assets are redistributed &
\citep{sattler2018benchmarking,zhang2021referencepose} \\
HLoc & Visual localization pipeline &
Apache-2.0 license for the HLoc codebase; third-party components follow their own licenses &
\citep{sarlin2019coarse} \\
JamMa codebase & Base implementation adapted for WAM &
JamMa code is released under MIT license. Its repository notes that parts of the code are based on upstream LoFTR-family repositories under their own licenses, terms, and conditions; we preserve applicable notices and do not redistribute third-party checkpoints &
\citep{lu2025jamma,sun21loftr} \\
Mamba / selective SSM & Reference implementation for WovenMamba SSM-style blocks &
Apache-2.0 license for the official Mamba implementation &
\citep{Gu2024Mamba} \\
ConvNeXtV2-style encoder & Backbone architecture and initialization &
MIT license for official ConvNeXtV2 code. Official ImageNet pre-trained/fine-tuned weights are under CC-BY-NC; we use them under those terms and do not redistribute third-party weights &
\citep{woo2023convnextv2} \\
GlueFactory / LightGlue & HPatches evaluation code and baseline evaluation utilities &
Apache-2.0 license for GlueFactory and LightGlue code/trained models; third-party models used through GlueFactory follow their own licenses &
\citep{sarlin23lightglue} \\
\bottomrule
\end{tabular}
\end{table}

For other baseline matchers, we cite the corresponding papers and follow the licenses and terms of their official implementations and checkpoints.

\clearpage
\section*{NeurIPS Paper Checklist}

\begin{enumerate}

\item {\bf Claims}
    \item[] Question: Do the main claims made in the abstract and introduction accurately reflect the paper's contributions and scope?
    \item[] Answer: \answerYes{}
    \item[] Justification: The abstract and introduction state the main scope and contributions: role-specialized intra-image and inter-image aggregation, WovenMamba, and AnchorCrossAttention. These claims are supported by the benchmark comparisons and controlled ablations in Sections~\ref{sec:benchmark_comparison}--\ref{sec:ablations}.

\item {\bf Limitations}
    \item[] Question: Does the paper discuss the limitations of the work performed by the authors?
    \item[] Answer: \answerYes{}
    \item[] Justification: The paper discusses limitations in Section~\ref{sec:discussion_and_limitations} and Appendix~\ref{app:broader_impact}, including the nearest-neighbor matching strategy, Aachen localization behavior, responsible-use considerations, and computational scaling of the context module. Appendix~\ref{app:resolution_scaling} provides a diagnostic resolution-scaling study supporting the computational discussion.

\item {\bf Theory assumptions and proofs}
    \item[] Question: For each theoretical result, does the paper provide the full set of assumptions and a complete (and correct) proof?
    \item[] Answer: \answerNA{}
    \item[] Justification: The paper does not present formal theoretical results or theorems. The receptive-field discussion in Section~\ref{sec:wovenmamba} is an illustrative simplified linear analysis, and the simplifying assumptions are stated there.

    \item {\bf Experimental result reproducibility}
    \item[] Question: Does the paper fully disclose all the information needed to reproduce the main experimental results of the paper to the extent that it affects the main claims and/or conclusions of the paper (regardless of whether the code and data are provided or not)?
    \item[] Answer: \answerYes{}
    \item[] Justification: The paper describes the model components, architecture hyperparameters, training setup, datasets, evaluation metrics, RANSAC settings, runtime protocol, and ablation protocol in Sections~\ref{sec:method}--\ref{sec:experiments} and Appendix~\ref{app:implementation}. These details provide a reasonable path to reproduce the main results.

\item {\bf Open access to data and code}
    \item[] Question: Does the paper provide open access to the data and code, with sufficient instructions to faithfully reproduce the main experimental results, as described in supplemental material?
    \item[] Answer: \answerYes{}
    \item[] Justification: Code and a pretrained checkpoint are publicly available at \url{https://github.com/mmaiLab/WovenAnchorMatcher}. The repository provides setup and data access instructions, MegaDepth training, evaluation on MegaDepth-1500, ScanNet-1500, and HPatches, and ablation configurations. The release focuses on WAM; baseline-specific scripts, the seven-run seed/checkpoint mapping, and Aachen/HLoc integration are not included. Dataset access remains subject to the original terms.

\item {\bf Experimental setting/details}
    \item[] Question: Does the paper specify all the training and test details (e.g., data splits, hyperparameters, how they were chosen, type of optimizer) necessary to understand the results?
    \item[] Answer: \answerYes{}
    \item[] Justification: Sections~\ref{sec:experimental_setup} and Appendix~\ref{app:implementation} specify the datasets, image resizing, optimizer, learning rate schedule, batch size, training duration, architecture hyperparameters, matching thresholds, RANSAC settings, and runtime measurement protocol.

\item {\bf Experiment statistical significance}
    \item[] Question: Does the paper report error bars suitably and correctly defined or other appropriate information about the statistical significance of the experiments?
    \item[] Answer: \answerNo{}
    \item[] Justification: Appendix~\ref{app:repeated_runs} reports mean, sample standard deviation, median, and min--max over seven WAM training runs on MegaDepth-1500. However, not all baselines and ablation variants are repeated due to computational cost. Ablation variants are controlled single-run comparisons unless otherwise stated.

\item {\bf Experiments compute resources}
    \item[] Question: For each experiment, does the paper provide sufficient information on the computer resources (type of compute workers, memory, time of execution) needed to reproduce the experiments?
    \item[] Answer: \answerYes{}
    \item[] Justification: Section~\ref{sec:experimental_setup} and Appendix~\ref{app:implementation} report the GPU type, peak allocated GPU memory, training time, batch size, and runtime protocol. Each WAM training run takes about 26 hours on one H100 GPU; the seven WAM runs used for variability analysis correspond to about 182 H100-GPU hours. Ablation variants use the same training schedule unless otherwise stated.
    
\item {\bf Code of ethics}
    \item[] Question: Does the research conducted in the paper conform, in every respect, with the NeurIPS Code of Ethics \url{https://neurips.cc/public/EthicsGuidelines}?
    \item[] Answer: \answerYes{}
    \item[] Justification: The work uses existing public computer-vision benchmarks and does not involve new human-subject data collection, crowdsourcing, or high-risk generative model release. We discuss responsible use and possible negative applications in Appendix~\ref{app:broader_impact}.

\item {\bf Broader impacts}
    \item[] Question: Does the paper discuss both potential positive societal impacts and negative societal impacts of the work performed?
    \item[] Answer: \answerYes{}
    \item[] Justification: Appendix~\ref{app:broader_impact} discusses positive applications such as localization, SfM, SLAM, mapping, image alignment, AR, and robotic perception, as well as negative uses such as privacy-invasive localization, surveillance, and reconstruction without consent.
    
\item {\bf Safeguards}
    \item[] Question: Does the paper describe safeguards that have been put in place for responsible release of data or models that have a high risk for misuse (e.g., pre-trained language models, image generators, or scraped datasets)?
    \item[] Answer: \answerNA{}
    \item[] Justification: The paper does not release high-risk generative models, scraped datasets, or assets that require special safeguards. It proposes a local feature matching method evaluated on existing public benchmarks.

\item {\bf Licenses for existing assets}
    \item[] Question: Are the creators or original owners of assets (e.g., code, data, models), used in the paper, properly credited and are the license and terms of use explicitly mentioned and properly respected?
    \item[] Answer: \answerYes{}
    \item[] Justification: The paper cites the datasets, benchmark pipelines, and model components used in the experiments. Appendix~\ref{app:broader_impact} lists their licenses or usage terms. The public repository preserves license texts and attribution for inherited code and documents the evaluation metadata separately. Dataset images, depth maps, and third-party checkpoints are obtained from their respective sources rather than bundled; ScanNet access remains subject to its Terms of Use.

\item {\bf New assets}
    \item[] Question: Are new assets introduced in the paper well documented and is the documentation provided alongside the assets?
    \item[] Answer: \answerYes{}
    \item[] Justification: The released code and pretrained checkpoint are documented in the repository README and setup guide, covering installation, data access, training, evaluation, and ablation usage. Original code contributions are licensed under MIT; inherited components retain their respective licenses. The pretrained checkpoint is released under CC BY-NC 4.0. No new dataset is introduced.

\item {\bf Crowdsourcing and research with human subjects}
    \item[] Question: For crowdsourcing experiments and research with human subjects, does the paper include the full text of instructions given to participants and screenshots, if applicable, as well as details about compensation (if any)? 
    \item[] Answer: \answerNA{}
    \item[] Justification: The work does not involve crowdsourcing, participant studies, or new human-subject data collection.

\item {\bf Institutional review board (IRB) approvals or equivalent for research with human subjects}
    \item[] Question: Does the paper describe potential risks incurred by study participants, whether such risks were disclosed to the subjects, and whether Institutional Review Board (IRB) approvals (or an equivalent approval/review based on the requirements of your country or institution) were obtained?
    \item[] Answer: \answerNA{}
    \item[] Justification: The work does not involve crowdsourcing or research with human subjects, so IRB approval or equivalent review is not applicable.

\item {\bf Declaration of LLM usage}
    \item[] Question: Does the paper describe the usage of LLMs if it is an important, original, or non-standard component of the core methods in this research? Note that if the LLM is used only for writing, editing, or formatting purposes and does \emph{not} impact the core methodology, scientific rigor, or originality of the research, declaration is not required.
    \item[] Answer: \answerNA{}
    \item[] Justification: The core method, experiments, and analysis do not use LLMs as an important, original, or non-standard component. Any writing or editing assistance, if used, does not affect the methodology, scientific rigor, or originality of the research.

\end{enumerate}

\end{document}